\documentclass[10pt,twocolumn,letterpaper]{article}

\usepackage{cvpr}              

\usepackage{float}
\usepackage{dblfloatfix}
\usepackage{caption}
\usepackage[most]{tcolorbox}
\usepackage{placeins}

\definecolor{cvprblue}{rgb}{0.21,0.49,0.74}
\usepackage[pagebackref,breaklinks,colorlinks,allcolors=cvprblue]{hyperref}

\usepackage{caption}
\usepackage{xcolor}

\def\paperID{*****}
\def\confName{CVPR}
\def\confYear{2026}

\title{FoodSense: A Multisensory Food Dataset and Benchmark for Predicting Taste, Smell, Texture, and Sound from Images}

\author{
  Sabab Ishraq$^{1}$\quad
  Aarushi Aarushi$^{2}$\quad
  Juncai Jiang$^{2}$\quad
  Chen Chen$^{3}$\\[4pt]
  $^{1}$College of Engineering and Computer Science, University of Central Florida, Orlando, FL, USA\\
  $^{2}$College of Business Administration, University of Central Florida, Orlando, FL, USA\\
  $^{3}$Institute of Artificial Intelligence, University of Central Florida, Orlando, FL, USA\\
{\tt\small sabab.ishraq@ucf.edu, aarushi.aarushi@ucf.edu, jcjiang@ucf.edu, chen.chen@ucf.edu}
}

\begin{document}
\maketitle

\begin{abstract}
Humans routinely infer taste, smell, texture, and even sound from food images---a phenomenon well studied in cognitive science. However, prior vision language research on food has focused primarily on recognition tasks such as meal identification, ingredient detection, and nutrition estimation. Image-based prediction of multisensory experience remains largely unexplored. 
We introduce FoodSense, a human-annotated dataset for cross-sensory inference containing 66,842 participant-image pairs across 2,987 unique food images. Each pair includes numeric ratings (1-5) and free-text descriptors for four sensory dimensions: taste, smell, texture, and sound. To enable models to both predict and explain sensory expectations, we expand short human annotations into image-grounded reasoning traces. A large language model generates visual justifications conditioned on the image, ratings, and descriptors.
Using these annotations, we train FoodSense-VL, a vision language benchmark model to produce both multisensory ratings and grounded explanations directly from food images. This work connects cognitive science findings on cross-sensory perception with modern instruction tuning for multimodal models and shows that many popular evaluation metrics are insufficient for visually sensory inference. The dataset, code, and models are publicly available at \url{https://i-sababishraq.github.io/foodsense-vl/}.
\end{abstract}

\section{Introduction}
\label{sec:intro}

Humans infer gustatory, olfactory, and tactile qualities of food from visual cues alone. Viewing food images elicits taste-specific neural activity in gustatory cortex~\cite{doi:10.1073/pnas.2010932118}. Visual features such as color, texture, and plating shape perceived taste and flavor~\cite{doi:10.1177/20416695221127325,https://doi.org/10.1002/mar.20829,MOTOKI2023104996}, suggesting that visual information can convey rich multisensory signals. However, current vision-language models (VLMs) have largely ignored such multisensory experience inference from images.

This gap is reflected in existing food VLM benchmarks, which primarily evaluate recognition tasks, such as meal identification, ingredient detection, and macronutrient estimation~\cite{ma2023food500capfinegrainedfood,hosseinian2025januaryfoodbenchmarkjfb,romerotapiador2025visionlanguagemodelsreadydietary}. Romero-Tapiador et al.~\cite{romerotapiador2025visionlanguagemodelsreadydietary} find that current models struggle when asked to predict sensory properties such as texture, taste, or smell, indicating that multisensory inference remains largely unexplored. \textbf{To our knowledge, no existing dataset or benchmark explicitly models cross-modal sensory cues—taste, smell, texture, and sound—in a unified, image-grounded framework.}

Addressing this gap is practically important because images increasingly mediate food-related decisions. Consumers often rely on images when ordering food online or browsing recipes, forming expectations about taste and texture before any physical interaction. These motivations are also consistent with research in marketing and cognitive science showing that visual cues strongly shape consumers’ expectations about product experiences. In digital and retail contexts, images shape consumer expectations, and inaccurate expectations about taste or texture can reduce satisfaction~\cite{SIMMONDS2017340}. Tools that can anticipate likely sensory experiences from images could therefore improve digital food interfaces and recommendation systems. Such capabilities may also benefit individuals with sensory impairments (e.g., anosmia or aging-related sensory decline) or clinical settings that require texture-modified diets.

To study this problem, we introduce FoodSense, a multisensory food dataset containing 2{,}987 images annotated by 8{,}382 human raters across four sensory dimensions: taste, smell, texture, and sound. Each image is associated with multiple annotations including numeric ratings and short textual descriptors, resulting in 66{,}842 annotated image--participant pairs. These annotations capture how people infer sensory properties from visual appearance and how they describe the visual evidence supporting those expectations.

However, scalar ratings and short descriptors alone provide limited supervision for models that must predict sensory outcomes and explain them. Image-grounded reasoning is required to link visual cues to sensory expectations. Collecting such reasoning from humans at scale is costly and difficult to standardize. We therefore design an expansion pipeline that converts human ratings and descriptors into richer image-grounded explanations. A domain-specialized model filters hallucinated content~\cite{zhu2025judgelmfinetunedlargelanguage}, producing higher-quality reasoning traces suitable for instruction-style training of VLMs~\cite{cheng2025domainadaptiveposttrainingmultimodallarge,dettmers2023qloraefficientfinetuningquantized}. Using this supervision, we fine-tune Gemma~3 27B~\cite{gemmateam2025gemma3technicalreport} with a two-stage QLoRA training strategy~\cite{dettmers2023qloraefficientfinetuningquantized}. The first stage learns grounded sensory prediction from human ratings and descriptors, while the second stage trains the model to generate structured explanations using the expanded reasoning data. We find that single-stage training leads to rating collapse, whereas the two-stage design preserves discriminative sensory prediction while enabling explanatory generation. In summary, our contributions are:
\begin{itemize}
    \item \textbf{FoodSense}, a multisensory food dataset covering taste, smell, texture, and sound (Section~\ref{sec:dataset_intro}).
    \item \textbf{An expansion pipeline} that converts short human anchors into image-grounded natural-language explanations, validated by a domain-specialized model to filter hallucinated content.
    \item \textbf{A two-stage fine-tuning framework} for VLMs (FoodSense-VL) that produces both multisensory ratings and grounded explanations from food images alone, outperforming both generalist and domain-adapted VLM baselines on overall Pearson correlation.
    \item \textbf{A comprehensive evaluation} of open-source and commercial VLMs, showing that low absolute error can mask poor sensory discrimination and that Pearson correlation provides a more informative metric for this task.
\end{itemize}

\section{Related Work}
\label{sec:related}

\subsection{Cross-Sensory Inference and Food Perception}

A large body of cognitive science and sensory marketing research demonstrates that visual cues systematically shape taste, flavor, and texture perception~\cite{10.1007/978-3-319-23222-5_40,doi:10.1177/20416695221127325}. 
Spence et al. Reviews of this literature document robust effects of visual attributes such as color, shape, and texture on perceived taste intensity and quality~\cite{MOTOKI2023104996}. Experimental studies further show that combinations  of visual features interact to influence consumer expectation and choice behavior~\cite{VANDERLAAN201546}. Neuroimaging evidence also supports this connection: viewing food images can evoke taste-quality-specific activity in gustatory insular cortex~\cite{doi:10.1073/pnas.2010932118}. Grounded cognition and predictive coding theories suggest that visual cues trigger mental simulations of sensory experience~\cite{annurev:/content/journals/10.1146/annurev.psych.59.103006.093639,Clark_2013}. When images generate inaccurate expectations about taste or texture, expectation disconfirmation can reduce satisfaction and trust~\cite{PIQUERASFISZMAN2015165}. 

Cross-modal correspondences provide a key mechanism linking visual cues to sensory expectations. Prior work documents systematic associations between visual properties and taste qualities: red/pink $\leftrightarrow$ sweetness, green/yellow $\leftrightarrow$ sourness; rounded shapes $\leftrightarrow$ sweetness, angular shapes $\leftrightarrow$ sourness~\cite{doi:10.1177/20416695221127325}. Other studies show that visual cues can alter perceived text or mouthfeel. For instance, color influences perceived texture~\cite{https://doi.org/10.1002/mar.20829}, while contextual visual information can modulate perceived taste intensity~\cite{10.3389/fcomp.2025.1512931,SUGIMORI2022104539}. Together, these regularities suggest that visual cues contain structured information about likely sensory experiences, motivating computational approaches that predict such expectations directly from images. 

Prior computational research on taste and flavor prediction has largely focused on molecular and graph-based approaches~~\cite{GUNNING2025101127}. While these approaches model chemical determinants of flavor, they do not address how humans infer sensory qualities directly from visual appearances.
{\textit{In contrast, our work investigates whether multisensory expectations can be predicted from food images alone. By introducing a novel, large-scale dataset of human sensory annotations (Section~\ref{sec:dataset_intro}), we provide the necessary foundation for end-to-end visual cross-sensory inference.}}

\subsection{Vision Language Models for Food}

Recent VLMs have been adapted to food domains through specialized benchmarks, datasets, and post-training. 
The January Food Benchmark~\cite{hosseinian2025januaryfoodbenchmarkjfb} introduces a 1K-image benchmark where january/food-vision-v1 achieves 86.2\% vs.\ 74.1\% for GPT-4o. 
AdaptLLM~\cite{cheng2025domainadaptiveposttrainingmultimodallarge} post-trains MLLMs on 131K food-visual instructions for recipe generation and ingredient identification. 
Food-500 Cap~\cite{ma2023food500capfinegrainedfood} evaluates VLMs on fine-grained food captions. 
FoodNExTDB~\cite{romerotapiador2025visionlanguagemodelsreadydietary} tests six VLMs on 9.2K expert-labeled images, finding that closed-source models exceed 90\% on single-product recognition but struggle with fine-grained cooking styles and textures.

Despite strong performance on these recognition-oriented tasks, existing VLMs remain limited in predicting fine-grained, cross-modal sensory experiences--such as taste, sound, smell, and texture--from visual input alone, as our evaluation in Section~\ref{sec:experiments} shows. \textit{Our work addresses this limitation by shifting the paradigm from recognition to multisensory prediction. We introduce a dataset and training framework that enable VLMs to infer sensory properties directly from food images while generating image-grounded explanations.}

\subsection{Sensory Datasets and Prior Prediction Work}

Several datasets include human sensory ratings, but none are designed to support VLM training for image-based multisensory inference. 
 
Food Folio~\cite{10.3389/fpsyg.2020.585044} provides perceptual ratings for 138 images across 17 attributes by 1,054 participants. 
SFOOD~\cite{xu2025sfoodmultimodalbenchmarkcomprehensive} combines RGB with hyperspectral imaging to study  sweetness prediction and conclude that RGB imagery alone is insufficient for that task.  
Matsunaga et al.~\cite{10.1007/978-3-319-23222-5_40} estimate taste and texture using recipe ingredients together with images rather than image-only prediction.

Chemical and molecular approaches similarly predict taste from molecular structure~\cite{virtuous2024,beer2024}, addressing a different modality than our image-based setting. To our knowledge, no prior work trains VLMs to jointly generate multisensory ratings and natural-language explanations from food images alone. Our dataset (Section~\ref{sec:dataset_intro}) and accompanying benchmarking methodology (Section~\ref{sec:method}) directly address this unexplored visual-to-sensory inference task.
\section{FoodSense: A Multisensory Food Dataset}
\label{sec:dataset_intro}

We introduce FoodSense, a dataset specifically designed to study visual cross-sensory inference---how humans predict the taste, smell, texture, and sound of food from visual appearance alone. Unlike existing food datasets that focus on recognition or structured nutritional targets~\cite{ma2023food500capfinegrainedfood}, our dataset grounds rich, granular sensory language directly to images. The dataset therefore enables models to learn mappings between visual cues and human expectations about multisensory food experiences.

\begin{figure*}[!t]
\vspace{-0.3cm}
\centering
\begin{minipage}[c]{0.2\linewidth}
\centering
\includegraphics[width=\linewidth]{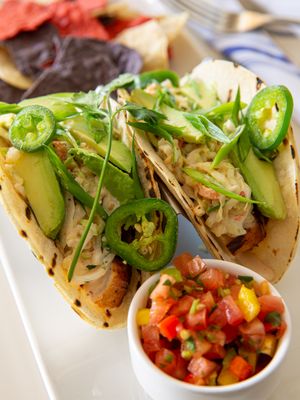}
\end{minipage}
\hfill
\begin{minipage}[c]{0.3\linewidth}
\centering
\scriptsize
\begin{tcolorbox}[colback=gray!5, colframe=gray!40,
  title={\scriptsize\textbf{Rating Task}},
  left=3pt, right=3pt, top=2pt, bottom=2pt, boxrule=0.5pt]
\textit{``Based on the image above, how would you rate the likely\ldots''}\\[2pt]
\begin{tabular}{lc}
\toprule
\textbf{Dimension} & \textbf{Scale} \\
\midrule
Taste   & 0 (Can't tell) / 1--7 \\
Smell   & 0 (Can't tell) / 1--7 \\
Texture & 0 (Can't tell) / 1--7 \\
Sound   & 0 (Can't tell) / 1--7 \\
\midrule
\multicolumn{2}{l}{1 = \textit{Very bad},\ \ 7 = \textit{Very good}} \\
\bottomrule
\end{tabular}\\[3pt]
\textit{``What do you think this food would sound like, taste like, smell like, and feel like (texture)?''}\\[2pt]
\textit{Please write \textbf{one or two words} for each sense.}
\end{tcolorbox}
\end{minipage}
\hfill
\begin{minipage}[c]{0.32\linewidth}
\centering
\scriptsize
\begin{tcolorbox}[colback=green!5, colframe=green!40,
  title={\scriptsize\textbf{Example Annotation (Taco)}},
  left=3pt, right=3pt, top=2pt, bottom=2pt, boxrule=0.5pt]
\begin{tabular}{lcc}
\toprule
\textbf{Sense} & \textbf{Rating (1--5)} & \textbf{Descriptor} \\
\midrule
Taste   & 4.2 & \textit{savory, spicy} \\
Smell   & 3.8 & \textit{smoky, fresh} \\
Texture & 4.5 & \textit{crispy, soft} \\
Sound   & 3.1 & \textit{crunchy} \\
\bottomrule
\end{tabular}\\[6pt]
\small Illustrative example. Ratings shown on the rescaled 1--5 scale; original responses collected on a 0--7 scale.
\end{tcolorbox}
\end{minipage}
\vspace{-0.3cm}
\caption{Annotation interface and example. \textbf{Left:} A food image as presented to participants (Taco, image 0005). \textbf{Center:} The structured rating task---participants rated each of four sensory dimensions on a 0--7 scale (0 = Can't tell from picture; 1 = Very bad; 7 = Very good) and provided one to two free-text descriptors per sense. \textbf{Right:} Illustrative annotation for a taco image showing rescaled ratings (1--5) and representative free-text descriptors across all four sensory dimensions.}
\label{fig:annotation_interface}
\end{figure*}

\subsection{Human Annotation}
\label{sec:data_collection}

 We started with the Yelp Open Dataset published for educational use (Source: \url{https://business.yelp.com/data/resources/open-dataset/}). The image pool reflects the cuisine diversity available on Yelp across the United States and is therefore broad, though not a controlled cross-cultural sample. We removed perceptual duplicates and filtered images containing identifiable faces. Then, through random sampling and manually verifying that each image depicts a single food item suitable for sensory evaluation, we obtained $N = 2{,}987$ structurally diverse food images. 

To generate ground-truth sensory labels, we administered a structured annotation survey to 8{,}382 participants recruited through a combination of an online panel and a university laboratory.Annotators were randomly assigned images via quota-based sampling in Qualtrics, making familiarity-based selection bias in assignment unlikely. Annotator-level demographic data was not collected. Each participant viewed one food image at a time and evaluated four sensory dimensions: taste, smell, texture, and sound. For each dimension, participants first rated their expected sensory experience using a seven-point Likert scale anchored at \textit{Very bad} (1) and \textit{Very good} (7). The survey prompt asked: \textit{``Based on the image above, how would you rate the likely\ldots''} followed by each sensory dimension. Participants could also select a \textit{Can't tell from picture} option (coded as 0) to flag visually ambiguous cases. To standardize the labels for model training, valid responses were linearly rescaled from the original 1--7 range to a 1--5 scale using $r_k = 1 + \frac{(r_{\text{orig}} - 1) \times 4}{6}$, while preserving relative ordering. Ratings marked as \textit{Can't tell from picture} were excluded from rescaling and retained as a separate binary \texttt{CanInfer}$_k$ flag. In addition to numeric ratings, participants provided one to two free-text words describing their expected sensory experience for each dimension---for example, \textit{crispy}, \textit{golden edges}, \textit{smoky}, or \textit{silent}. This dual-format design, combining structured ratings with natural language descriptors, $d_k$, captures both the magnitude of predicted sensory experience and the language people use to ground these judgments in visual evidence~\cite{VANDERLAAN201546}. Figure~\ref{fig:annotation_interface} illustrates the annotation interface and an example annotation for a taco image.

The initial annotation effort yielded 66,842 participant--image assessments across 2,987 images. During model training, 72 images were excluded due to filename inconsistencies in the training pipeline, reducing the pool to 65,348 annotations across 2,915 images. The complete dataset of 2,987 images is released publicly to support 
future research. Additional details on the annotation protocol, 
participant recruitment, and dataset statistics are provided in the 
Supplementary file~\ref{sec:suppl_annotation}--\ref{sec:suppl_stats}.

\begin{figure*}[!t]
\vspace{-0.3cm}
    \centering
    \includegraphics[width=0.85\textwidth]{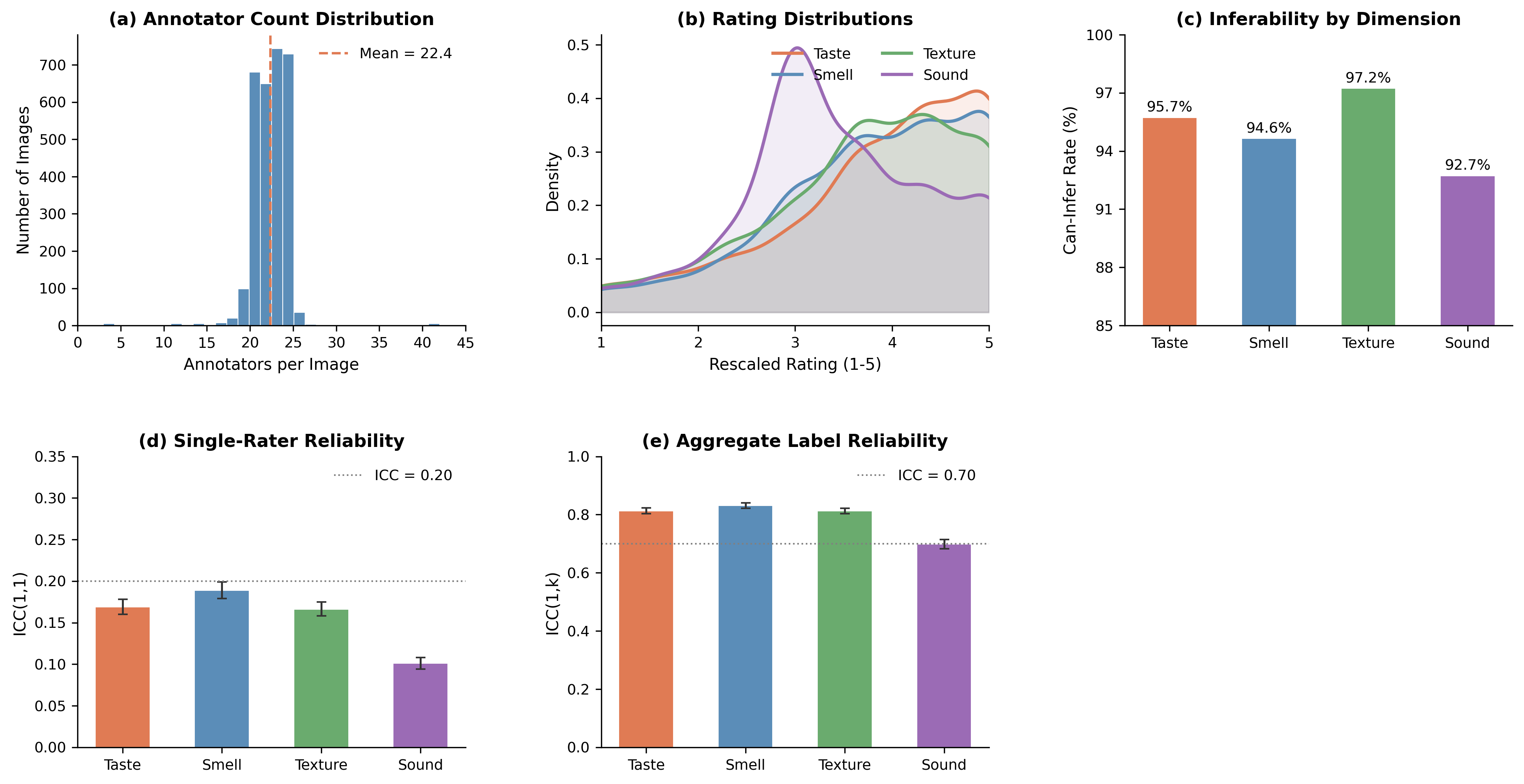}
    \vspace{-0.3cm}
    \caption{Dataset statistics for the Multisensory Food Dataset. (a)~Distribution of annotator counts per image (mean~$=22.38$, SD~$=2.02$). (b)~Kernel density estimates of rescaled ratings (1--5) per sensory dimension, restricted to \texttt{CanInfer}~$=1$ ratings. (c)~Proportion of ratings where participants could infer each sensory property from the image. (d)~Single-rater reliability ICC(1,1) with 95\% confidence intervals; dotted line marks ICC~$=0.20$. (e)~Aggregate label reliability ICC(1,k) with 95\% confidence intervals for mean ratings across $\bar{k}\approx21$ annotators; dotted line marks ICC~$=0.70$.}
    \label{fig:dataset_stats}
    \vspace{-0.5cm}
\end{figure*}
\subsection{Dataset Statistics and Reliability}
\label{sec:dataset_stats}
Figure~\ref{fig:dataset_stats} and Table~\ref{tab:dataset_stats} summarize the dataset statistics and inter-rater reliability. Each of the 2,987 images received an average of 22.38 annotations (SD~$=2.02$; range: 3--42). Can-Infer rates are high across all four sensory dimensions (92.7\%--97.2\%), indicating that participants generally considered food images sufficient for sensory inference. Sound is the least inferable dimension (92.7\%), which is consistent with the inherent difficulty of auditory inference from static images. The distributions of rescaled ratings (Fig.~\ref{fig:dataset_stats}b) are left-skewed across all dimensions, with means ranging from 3.44 (sound) to 3.89 (taste) on the rescaled 1--5 scale. This pattern suggests that participants generally anticipated positive sensory experiences from the foods shown in the images.
\begin{table}[!htbp]
\vspace{-0.3cm}
\centering
\caption{Descriptive statistics and inter-rater reliability for the full released dataset (2,987 images), computed on rescaled sensory ratings (1--5) with \texttt{CanInfer}~$=1$ per dimension. ICC(1,1): single-rater reliability; ICC(1,k): aggregate label reliability for mean of $\bar{k}$ annotators per image. All ICC values significant at $p<.0001$.}
\vspace{-0.2cm}
\label{tab:dataset_stats}
\resizebox{\columnwidth}{!}{%
\begin{tabular}{@{}lcccccc@{}}
\toprule
Dimension & $N$ & Mean & SD & ICC(1,1) [95\% CI] & ICC(1,k) [95\% CI] & $\bar{k}$ \\
\midrule
Taste   & 63,978 & 3.89 & 1.08 & 0.169~[0.160,~0.178] & 0.813~[0.803,~0.823] & 21.42 \\
Smell   & 63,259 & 3.83 & 1.05 & 0.189~[0.179,~0.199] & 0.831~[0.822,~0.840] & 21.18 \\
Texture & 64,986 & 3.74 & 1.06 & 0.166~[0.158,~0.175] & 0.813~[0.803,~0.822] & 21.76 \\
Sound   & 61,972 & 3.44 & 0.99 & 0.101~[0.094,~0.108] & 0.699~[0.683,~0.714] & 20.75 \\
\bottomrule
\end{tabular}%
}
\vspace{-0.3cm}
\end{table}
Inter-rater reliability was assessed using one-way intraclass correlation coefficients (ICCs)~\cite{shrout1979intraclass}, which quantify rating consistency under a random-raters assumption. ICC(1,1) measures single-rater consistency, indicating how well any individual annotator's ratings align with those of others. ICC(1,k) measures aggregate reliability for the mean rating across $\bar{k}$ annotators per image. Because model training uses per-image mean ratings rather than individual annotations, ICC(1,k) represents the operative reliability metric. Single-rater reliability ICC(1,1) ranges from 0.101 to 0.189 across the four sensory dimensions (Table~\ref{tab:dataset_stats}), reflecting the expected variability in subjective sensory inference. However, when aggregated across annotations per image ($\bar{k} \approx 21$), reliability increases substantially: ICC(1,k) ranges from 0.699 to 0.831. These values indicate good to excellent reliability of the image-level ground truth labels used for model training and evaluation. All ICC values are statistically significant ($p < .0001$). Annotation examples from the test set are provided in 
Appendix~\ref{sec:suppl_examples}. Descriptor-level consistency analyses provide converging evidence: 
per-image type-token ratios and top-$k$ descriptor coverage are 
reported in Appendix~\ref{sec:suppl_desc_consistency}, along with 
evidence that numeric rating disagreement and lexical diversity covary 
across images ($r = 0.249$ for taste; $r = 0.141$ for texture), 
validating both annotation signals. 
\section{Method}
\label{sec:method}

\subsection{Problem Statement}
\label{sec:problem_statement}

We formalize cross-modal sensory inference as a dense visual grounding problem. Given a food image $\mathcal{I}_i$, our objective is to predict a comprehensive sensory profile $\mathcal{Y}_i = \{ y_{i,\text{taste}}, y_{i,\text{smell}}, y_{i,\text{texture}}, y_{i,\text{sound}} \}$. To capture both subjective intensity and qualitative experience, each sensory dimension $k$ is modeled as a tuple $y_{i,k} = (\hat{r}_{i,k}, \hat{d}_{i,k}, \hat{p}_{i,k})$, where $\hat{r}_{i,k} \in [1,5]$ is a continuous scalar rating representing human-annotated sensory intensity, $\hat{d}_{i,k}$ is a discrete categorical anchor (a concise human descriptor), and $\hat{p}_{i,k}$ is an image-grounded rationale. The rationale $\hat{p}_{i,k}$ explicitly verbalizes the visual evidence within $\mathcal{I}_i$ (e.g., surface sheen, crumb structure, color gradients~\cite{FLEMING201462,annurev:/content/journals/10.1146/annurev-vision-102016-061429}) that justifies the predicted rating and descriptor outcomes.

Our multi-rater supervision in the FoodSense dataset ensures multiple participant annotations per image. Following Section~\ref{sec:data_curation}, we train and evaluate models on FoodSense using strict image-level isolation to prevent cross-contamination. Solving this formulation presents two fundamental challenges for standard auto-regressive VLMs:

\textbf{1. The Semantic Gap in Visual Grounding.} The FoodSense dataset provides highly structured and compact perceptual anchors $(\hat{r}_{i,k}, \hat{d}_{i,k})$, which precisely quantify human sensory expectations. However, mapping high-dimensional visual inputs directly to these compact text-scalar outputs presents a severe semantic bottleneck for generative VLMs~\cite{wang2024learningvisualgroundinggenerative}, which learn visual alignment optimally through dense textual dependencies. Direct fine-tuning on highly constrained labels causes models to memorize distribution statistics rather than learn causal visual grounding.

\textbf{2. Objective Conflict (Regression vs. Generation).} Simultaneously optimizing a VLM to predict an exact continuous scalar $\hat{r}_{i,k}$ (a regression-equivalent objective) alongside open-form explanatory text $\hat{p}_{i,k}$ introduces significant objective conflict, an issue documented in multi-task VLM architectures~\cite{xue2025regressioneovlmschallenge}. Gradient interference between highly constrained numeric prediction and unconstrained language modeling frequently causes representation collapse, where the model abandons discriminative capability and simply predicts an arbitrary dataset mean for all inputs.

\subsection{Method Overview}
To overcome these structural bottlenecks while honoring the precision of our human annotations, our framework introduces FoodSense-VL, a specialized training paradigm comprised of two components. 

First, to bridge the semantic gap in FoodSense annotations, we introduce an \textbf{Image-Grounded Expansion Framework} (Sec.~\ref{sec:expansion_framework}) inspired by MAmmoTH-VL~\cite{guo-etal-2025-mammoth}, which uses open Multimodal Large Language Models (MLLMs) to expand short anchors into richer rationales. Our framework utilizes an advanced zero-shot VLM to adapt the anchor$\rightarrow$expansion pattern to validate and expand our precise, compact human anchors $(\hat{r}_{i,k}, \hat{d}_{i,k})$ into the dense rationales $\hat{p}_{i,k}$ required for instruction-style supervision. 

Second, to mitigate objective conflict, we implement a \textbf{Two-Stage QLoRA Fine-Tuning Strategy} to train FoodSense-VL (Sec.~\ref{sec:training}), which decouples scalar grounding from rationale generation, thus avoiding the rating collapse observed in single-stage training.

\subsection{Data Curation and Splits}
\label{sec:data_curation}

Validating visual-sensory inference for VLMs requires strict image-level isolation to ensure that models learn generalizable cross-sensory mappings rather than memorizing specific food images. To construct the partitions, we utilize a pseudo-random stratified shuffle split at the image level (75\% / 10\% / 15\% for train / validation / test) based on the binned mean overall rating of each image. While this stratification mitigates distributional biases by ensuring the partitions share identical sensory rating distributions, we acknowledge that completely removing cultural biases is infeasible because the participants are demographically accustomed to the depicted foods. However, given that the United States encompasses a highly diverse mix of demographics and widespread consumption of popular global cuisines, the resulting collection serves as a satisfactorily diverse dataset. To train FoodSense-VL, we retain assessments where participants indicated visual inferability for all four sensory dimensions (\texttt{CanInfer}$_k = 1$) only, resulting in the final training dataset of $N = 2{,}915$ food images with 58,443 participant--image annotations (approximately 20 per image). Ultimately, the stratification assigns the 58,443 annotations into strict, non-overlapping partitions: training (43,758 ratings across 2,185 images), validation (5,834 ratings across 292 images), and test (8,851 ratings across exactly 438 images).

To support instruction-style VLMs training, we synthesize aggregated human ratings ($\bar{r}_k$) and descriptors ($d_k^*$) into natural, conversational-style rationales. Specifically, we apply a targeted expansion pipeline~\cite{guo-etal-2025-mammoth} that maps these human anchors into image-grounded text, then validate outputs with a specialized judge model to filter hallucinated content~\cite{zhu2025judgelmfinetunedlargelanguage,cheng2025domainadaptiveposttrainingmultimodallarge}.

\subsection{Image-Grounded Expansion Framework}
\label{sec:expansion_framework}
While the FoodSense dataset provides highly reliable scalar ratings and short descriptors, direct end-to-end training on these sparse labels in FoodSense is insufficient for teaching compositional visual reasoning~\cite{romerotapiador2025visionlanguagemodelsreadydietary}. To address this, we introduce a structured distillation pipeline that expands these compact perceptual annotations into dense, image-grounded reasoning traces suitable for fine-tuning stage. The expansion framework consists of several core phases:

\noindent\textbf{1. Establishing the Visual-Sensory Anchor.} To capture the ground-truth perceptual consensus for a given image $\mathcal{I}_i$, we first compute the aggregate sensory anchor across $M$ participants. For each sensory dimension $k$, we compute the mean intensity rating $\bar{r}_k$ and identify the primary semantic descriptor $d_k^*$ (the first valid non-empty annotation):
\begin{align}
\bar{r}_k &= \frac{1}{M} \sum_{m=1}^M r_{k,m}, \label{eq:mean_rating} \\
d_k^* &= d_{k,m^*}, \quad m^* = \min \{ m \mid d_{k,m} \neq \emptyset \} \label{eq:desc_select}
\end{align}
\vspace{-0.3cm}

\noindent\textbf{2. Image-Grounded Rationale Expansion.} We employ a state-of-the-art Multimodal Large Language Model (Gemma 3 27B IT) as a teacher model to perform zero-shot visual reasoning. Conditioned on both the high-dimensional image features of $\mathcal{I}_i$ and the strict constraints of the aggregate anchor $\{\bar{r}_k, d_k^*\}$, the teacher synthesizes a dense rationale $p_k$:

\begin{equation}
p_k = \mathcal{T}_{\theta}\bigl(\mathcal{I}_i, \textsc{Prompt}(\bar{r}_k, d_k^*)\bigr) \label{eq:expansion}
\end{equation}

The system prompt casts the teacher as a sensory analysis expert, explicitly instructing it to scan the visual context and generate two to three sentences explaining exactly what specific features within the image (e.g., surface topology, ingredient visibility) support the human-annotated anchor, while enforcing diverse, non-templated sentence structures.

\noindent\textbf{3. Multimodal Hallucination Filtering (VLM-as-Judge):} To prevent the propagation of generative hallucinations into the training corpus, we introduce an independent vision-language judge (AdaptLLM food-Llama 11B). The judge evaluates whether full sensory blocks $[\bar{r}_k, d_k^*, p_k]$ logically match the image and avoid hallucinating visual details not actually present. Any expansion returning ``NO'' (or lacking a clear ``YES'') is categorically rejected and discarded, forcing the downstream training pipeline to fall back on the original sparse human descriptor for that instance.

\noindent\textbf{4. Rationale Parsing:} The model is prompted to output four lines of the form ``Sense (X.X/5.0): descriptor. [expansion].'' We use regex to isolate each sense block and split it at the first ``.'' into descriptor and expansion; only the expansion is stored. If the split fails (e.g., no period), the full block is kept.

\noindent\textbf{5. Output.} The validated rationales are written to a JSON file $\mathcal{I}_i \mapsto \{ \text{taste}: p_t, \text{smell}: p_s, \text{texture}: p_\tau, \text{sound}: p_o \}$, which supplies the dense supervision targets for fine-tuning.

\subsection{Stage-Two Training}
\label{sec:training}

We employ two-stage fine-tuning to produce outputs of the form:

\noindent\textbf{Stage 1: Scalar and descriptor grounding.}
In the first stage, we train the model solely on per-participant ratings and descriptors \((r_{i,k,m}, d_{i,k,m})\), without any rationale text. The loss reduces to a language modeling loss focused on accurately emitting the scalar scores and short descriptors, conditioned on the image and a standardized prompt template. This stage encourages the model to align its visual representations with human sensory anchors before reasoning text is introduced.

\noindent\textbf{Stage 2: Rationale generation on calibrated backbone.}
In the second stage, we initialize from the Stage~1 checkpoint and introduce per-image rationales \(\hat{p}_{i,k}\) obtained from the expansion framework (Sec.~\ref{sec:expansion_framework}). The model is now trained to jointly produce ratings, descriptors, and rationales, but the scalar grounding head and prompted format are kept fixed; LoRA adapters continue to update the shared representation layers. In this stage the model preserves its calibrated rating behavior while learning to emit rich, image-grounded explanations.

This combination of image-grounded rationale distillation and staged low-rank adaptation yields FoodSense-VL, a VLM that can (i) predict calibrated multisensory ratings and descriptors and (ii) justify them with visually faithful, domain-specific rationales, aligning with recent findings that rich rationales significantly enhance multimodal reasoning~\cite{guo-etal-2025-mammoth}.

\definecolor{ourscolor}{RGB}{230,245,255}

\begin{figure}[t]
\vspace{-0.3cm}
\centering
\footnotesize
\setlength{\parskip}{0pt}
\setlength{\lineskip}{0.5pt}

\begin{minipage}[c]{0.18\columnwidth}
\centering
\includegraphics[width=\linewidth]{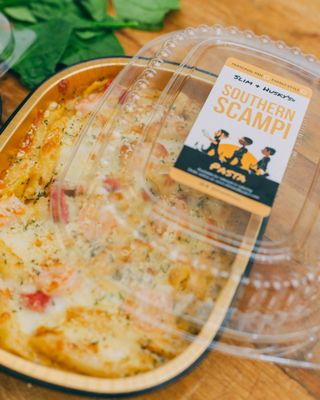}
\end{minipage}%
\hfill $\rightarrow$ \hfill
\begin{minipage}[c]{0.22\columnwidth}
\begin{tcolorbox}[colback=gray!5, colframe=gray!40,
  title={\tiny\textbf{Human}},
  left=1pt, right=1pt, top=0.5pt, bottom=0.5pt, boxrule=0.3pt]
\tiny
\begin{tabular}{@{}lc@{}}
Taste   & 4.0 \\
Smell   & 3.6 \\
Texture & 3.2 \\
Sound   & 2.9 \\
\end{tabular}
\end{tcolorbox}
\end{minipage}%
\hfill $\rightarrow$ \hfill
\begin{minipage}[c]{0.12\columnwidth}
\centering
\includegraphics[width=1.0\linewidth]{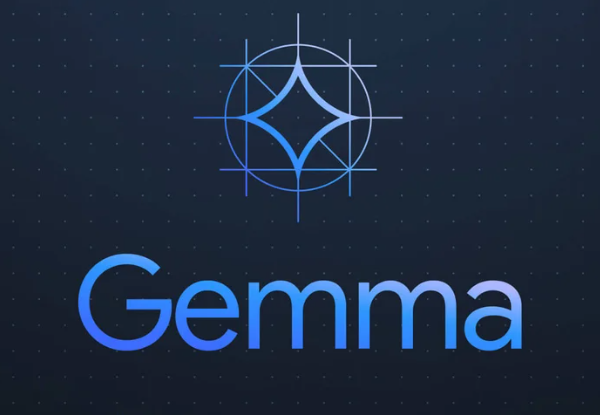}\\
\tiny Gemma
\end{minipage}%
\hfill $\rightarrow$ \hfill
\begin{minipage}[c]{0.32\columnwidth}
\begin{tcolorbox}[colback=cyan!5, colframe=cyan!40,
  title={\tiny\textbf{Rationale}},
  left=1pt, right=1pt, top=0.5pt, bottom=0.5pt, boxrule=0.3pt]
\tiny
\textit{Texture:} crunchy. Crisp tortilla, cabbage, salsa suggest satisfying crunch.
\end{tcolorbox}
\end{minipage}

\vspace{1pt}
\makebox[\columnwidth][r]{\makebox[0.12\columnwidth][c]{$\downarrow$}}
\vspace{1pt}
\begin{minipage}[c]{0.55\columnwidth}
\begin{tcolorbox}[colback=green!5, colframe=green!40,
  title={\tiny\textbf{Output}},
  left=1pt, right=1pt, top=0.5pt, bottom=0.5pt, boxrule=0.3pt]
\tiny
3.2. \textit{Texture:} Soft. Bubbling cheese and pasta suggest soft, yielding texture. No crispy elements; predominantly smooth, creamy mouthfeel.
\end{tcolorbox}
\end{minipage}%
\hfill $\leftarrow$ \hfill
\begin{minipage}[c]{0.12\columnwidth}
\centering
\includegraphics[width=1.0\linewidth]{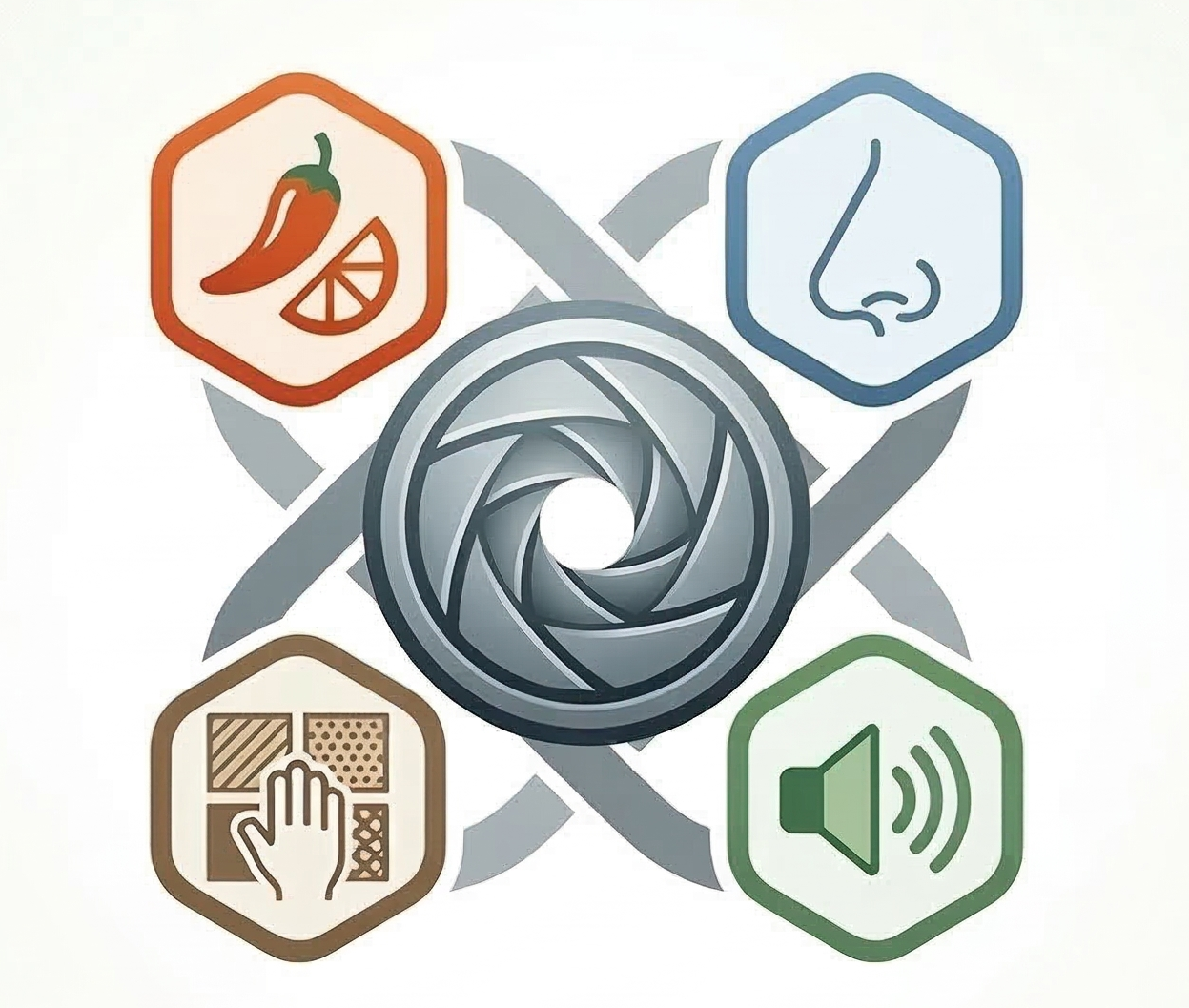}\\
\tiny FoodSense-VL
\end{minipage}%
\hfill $\leftarrow$ \hfill
\begin{minipage}[c]{0.12\columnwidth}
\centering
\includegraphics[width=1.0\linewidth]{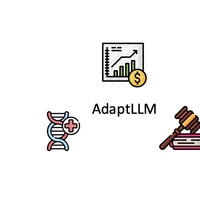}\\
\tiny Judge
\end{minipage}

\vspace{-0.2cm}
\caption{\textbf{Pipeline.} A Southern Scampi image with human sensory annotations is expanded by Gemma~3 27B IT into image-grounded rationales; Food-Llama judges and filters hallucinated content. FoodSense-VL predicts ratings and explanations from images alone. The output box shows an example texture prediction with visual justification.}
\label{fig:pipeline}
\vspace{-0.4cm}
\end{figure}

\section{Experiments}
\label{sec:experiments}

To systematically evaluate the capability of VLMs in predicting non-visual properties from images, our evaluation is guided by four core Research Questions (RQs):
\begin{itemize}
    \item \textbf{RQ1 (Feasibility):} To what extent can VLM infer complex cross-sensory properties (taste, smell, texture, sound) from visual cues alone?
    \item \textbf{RQ2 (Effectiveness):} Does domain-specific fine-tuning on human-annotated sensory data outperform state-of-the-art generalist and food-specific VLMs?
    \item \textbf{RQ3 (Sensory Variation):} How do models perform across different sensory dimensions, and which properties are the most challenging to visually infer?
    \item \textbf{RQ4 (Interpretability):} Can FoodSense-VL, our two-stage tuned model, generate grounded, natural-language rationales that align with its quantitative ratings?
\end{itemize}

\noindent \textbf{Experimental Setup.}
We benchmark FoodSense-VL against InternVL2.5-26B~\cite{chen2024internvlscalingvisionfoundation}, LLaVA-v1.6-34B~\cite{liu2023visualinstructiontuning}, Qwen2.5-VL-32B~\cite{qwen2025qwen25technicalreport},and the domain-specific Food-LLaMA-11B~\cite{cheng2025domainadaptiveposttrainingmultimodallarge}.

FoodSense-VL is built on the 4-bit QLoRA~\cite{dettmers2023qloraefficientfinetuningquantized} Gemma~3 27B IT~\cite{gemmateam2025gemma3technicalreport} architecture, on a single NVIDIA H100 GPU provided via the ACCESS program~\cite{10.1145/3569951.3597559}. In Stage~1 (Human Grounding), we train with a learning rate of $5\times10^{-6}$.  Stage~1 (human grounding) aligns the model to the human evaluations using a learning rate of lr=$5\times10^{-6}$; Stage~2 (reasoning integration) resumes from Stage~1 along with the expansion rationales using learning rate of$2\times10^{-6}$. Both stages use effective batch size 64 and cosine scheduling.

\begin{tcolorbox}[colback=gray!5, colframe=gray!50, title={\scriptsize\textbf{Evaluation Prompt}}, left=4pt, right=4pt, top=2pt, bottom=2pt, boxrule=0.5pt]
\footnotesize\texttt{Analyze this food image and estimate the likely sensory experience. Plain text only (no markdown; no bullets; no asterisks). Provide a Sensory Assessment with 4 sections, one each for Taste, Smell, Texture, and Sound, formatted as: Sense (X.X/5.0): 3-4 sentences of detailed visual justification.}
\end{tcolorbox}

\noindent \textbf{Evaluation Metrics.}
Ratings $\hat{r}_{k,i}$ are parsed via regex and compared to human means $r_{k,i}^*$ ($N=438$ images, $K=4$ senses) via the following metrics: 
\begin{align}
\small
r &= \frac{\sum_i(\hat{r}_i-\bar{\hat{r}})(r_i^*-\bar{r}^*)}{\sqrt{\sum_i(\hat{r}_i-\bar{\hat{r}})^2}\;\sqrt{\sum_i(r_i^*-\bar{r}^*)^2}} \label{eq:pearson}\\
\rho_c &= \frac{2r\,\sigma_{\hat{r}}\sigma_{r^*}}{\sigma_{\hat{r}}^2 + \sigma_{r^*}^2 + (\mu_{\hat{r}}-\mu_{r^*})^2} \label{eq:ccc}
\end{align}
We report MAE and RMSE as absolute-error metrics (lower is better): MAE reflects average deviation from human means, while RMSE penalizes larger errors more strongly. We also report Pearson~$r$~\cite{a0dc553c-0830-3fe2-ab2a-eece0d66a7db}, Spearman~$\rho$~\cite{10.1093/ije/dyq191}, Lin's Concordance Correlation Coefficient (CCC)~\cite{ae7889f8-f8e6-35c0-919b-64c18f0b9197}, and Ordinal Accuracy (3-class: Low/Med/High).

\subsection{Quantitative Results}

Tables~\ref{tab:sensory_mae}--\ref{tab:sensory_pearson} compare our two-stage model against open-source VLMs and the untuned Gemma~3 base. We report per-sense and overall results for MAE~(Table~\ref{tab:sensory_mae}), Spearman~$\rho$ and Lin's CCC~(Table~\ref{tab:sensory_rank}), and Pearson~$r$ with Ordinal Accuracy~(Table~\ref{tab:sensory_pearson}). Human inter-rater MAE is $1.04$ on average, placing all models well below the human disagreement ceiling.

\begin{table}[!htbp]
\centering
\caption{MAE$\downarrow$ and RMSE$\downarrow$ by sensory dimension.} 
\vspace{-0.3cm}
\label{tab:sensory_mae}
\renewcommand{\arraystretch}{1.4}
\resizebox{\columnwidth}{!}{%
\begin{tabular}{@{}lcccccccccc@{}}
\toprule
 & \multicolumn{5}{c}{MAE$\downarrow$} & \multicolumn{5}{c}{RMSE$\downarrow$} \\
\cmidrule(lr){2-6}\cmidrule(lr){7-11}
Model & Taste & Smell & Tex. & Sound & Avg & Taste & Smell & Tex. & Sound & Avg \\
\midrule
LLaVA-v1.6-34B         & 0.601 & 0.426 & \textbf{0.406} & \textbf{0.304} & \textbf{0.435} & 0.745 & 0.549 & \textbf{0.542} & \textbf{0.389} & \textbf{0.556} \\
InternVL2.5-26B        & 0.635 & \textbf{0.423} & 0.663 & 0.305 & 0.507 & 0.787 & \textbf{0.543} & 0.802 & 0.390 & 0.630 \\
Qwen2.5-VL-32B         & 0.528 & 0.432 & 0.667 & 0.730 & 0.589 & 0.673 & 0.569 & 0.808 & 0.850 & 0.725 \\
Gemma~3 27B (base)     & \textbf{0.411} & 0.505 & 0.511 & 0.980 & 0.602 & \textbf{0.504} & 0.617 & 0.633 & 1.067 & 0.705 \\
Food-LLaMA-11B         & 0.558 & 0.764 & 0.532 & 1.108 & 0.740 & 0.706 & 0.979 & 0.683 & 1.337 & 0.926 \\
\textbf{FoodSense-VL (Ours)} & 0.515 & 0.525 & 0.531 & 0.581 & 0.538 & 0.645 & 0.660 & 0.678 & 0.723 & 0.676 \\
\bottomrule
\end{tabular}%
}
\end{table}

\begin{table}[!htbp]
\centering
\vspace{-0.1cm}
\caption{Spearman $\rho\uparrow$ and Lin's Concordance Correlation Coefficient (CCC)$\uparrow$ by sensory dimension. Best in \textbf{bold}.}
\vspace{-0.3cm}
\label{tab:sensory_rank}
\renewcommand{\arraystretch}{1.4}
\resizebox{\columnwidth}{!}{%
\begin{tabular}{@{}lcccccccccc@{}}
\toprule
 & \multicolumn{5}{c}{Spearman $\rho\uparrow$} & \multicolumn{5}{c}{Lin's CCC$\uparrow$} \\
\cmidrule(lr){2-6}\cmidrule(lr){7-11}
Model & Taste & Smell & Tex. & Sound & Avg & Taste & Smell & Tex. & Sound & Avg\\
\midrule
InternVL2.5-26B        & 0.171 & 0.171 & 0.227 & 0.140 & 0.177 & 0.028 & 0.063 & 0.086 & 0.134 & 0.078 \\
LLaVA-v1.6-34B         & 0.271 & 0.147 & 0.189 & 0.181 & 0.197 & 0.115 & 0.064 & 0.113 & 0.159 & 0.113 \\
Qwen2.5-VL-32B         & 0.311 & 0.141 & 0.225 & \textbf{0.265} & 0.236 & 0.159 & 0.121 & 0.078 & 0.140 & 0.124 \\
Gemma~3 27B (base)     & 0.196 & 0.178 & 0.140 & 0.208 & 0.181 & 0.159 & 0.151 & 0.181 & 0.053 & 0.136 \\
Food-LLaMA-11B         & 0.085 & 0.036 & 0.108 & 0.127 & 0.089 & 0.074 & 0.010 & 0.097 & 0.040 & 0.055 \\
\textbf{FoodSense-VL (Ours)} & \textbf{0.395} & \textbf{0.398} & \textbf{0.395} & 0.253 & \textbf{0.360} & \textbf{0.381} & \textbf{0.382} & \textbf{0.372} & \textbf{0.236} & \textbf{0.343} \\
\bottomrule
\end{tabular}%
}
\end{table}
\vspace{-0.5cm}

\begin{table}[!htbp]
\centering
\caption{Pearson $r\uparrow$ and Ordinal Accuracy$\uparrow$ (3-class: Low/Med/High) by sensory dimension. Best in \textbf{bold}.}
\vspace{-0.3cm}
\label{tab:sensory_pearson}
\renewcommand{\arraystretch}{1.4}
\resizebox{\columnwidth}{!}{%
\begin{tabular}{@{}lcccccccccc@{}}
\toprule
 & \multicolumn{5}{c}{Pearson $r\uparrow$} & \multicolumn{5}{c}{Ordinal Accuracy$\uparrow$} \\
\cmidrule(lr){2-6}\cmidrule(lr){7-11}
Model & Taste & Smell & Tex. & Sound & Avg & Taste & Smell & Tex. & Sound & Avg \\
\midrule
InternVL2.5-26B        & 0.229 & 0.226 & 0.262 & 0.187 & 0.226 & 0.694 & 0.646 & 0.600 & 0.719 & 0.665 \\
LLaVA-v1.6-34B         & 0.328 & 0.143 & 0.234 & 0.208 & 0.229 & \textbf{0.721} & 0.651 & 0.612 & \textbf{0.726} & \textbf{0.678} \\
Qwen2.5-VL-32B         & 0.312 & 0.158 & 0.233 & 0.280 & 0.246 & 0.703 & \textbf{0.671} & 0.605 & 0.616 & 0.649 \\
Gemma~3 27B (base)     & 0.217 & 0.178 & 0.184 & 0.263 & 0.211 & 0.696 & 0.489 & 0.564 & 0.600 & 0.587 \\
Food-LLaMA-11B         & 0.075 & 0.013 & 0.123 & 0.109 & 0.080 & 0.561 & 0.415 & 0.618 & 0.300 & 0.474 \\
\textbf{FoodSense-VL (Ours)} & \textbf{0.398} & \textbf{0.401} & \textbf{0.397} & \textbf{0.290} & \textbf{0.372} & 0.705 & 0.662 & \textbf{0.642} & 0.660 & 0.667 \\
\bottomrule
\end{tabular}%
}
\end{table}

\begin{table}[!htbp]
\centering
\caption{Ablation: Single-stage (flat) vs.\ two-stage (curriculum) fine-tuning.}
\vspace{-0.3cm}
\label{tab:ablation}
\renewcommand{\arraystretch}{1.3}
\resizebox{\columnwidth}{!}{%
\begin{tabular}{@{}lccccccc@{}}
\toprule
Setting & MAE$\downarrow$ & Pearson$\uparrow$ & Spearman$\uparrow$ & CCC$\uparrow$ & $\sigma_{\text{pred}}$ & Ord.Acc$\uparrow$ & W/in 1$\uparrow$ \\
\midrule
Single-Stage   & \textbf{0.589} & 0.254 & 0.227 & 0.169 & 0.367 & \textbf{0.593} & \textbf{0.834} \\
Two-Stage (FoodSense-VL)              & 0.633 & \textbf{0.297} & \textbf{0.274} & \textbf{0.238} & \textbf{0.591} & 0.542 & 0.793 \\
\midrule
$\Delta$ (Two$-$Single) & {\color{red}+0.044} & {\color{green!50!black}+0.043} & {\color{green!50!black}+0.047} & {\color{green!50!black}+0.069} & {\color{green!50!black}+0.224} & {\color{red}$-$0.051} & {\color{red}$-$0.041} \\
\bottomrule
\end{tabular}%
}
\end{table}
\begin{figure}[!t]
\vspace{-0.2cm}
\centering
\setlength{\parskip}{0pt}
\setlength{\lineskip}{1pt}

\includegraphics[width=0.62\columnwidth]{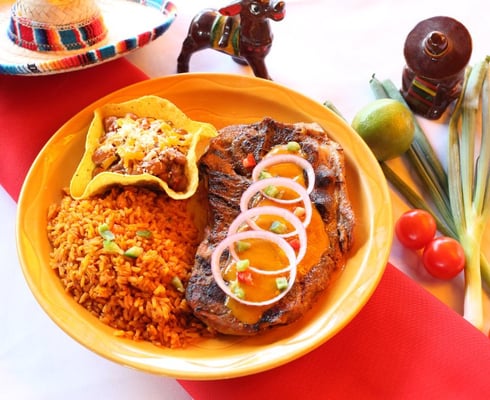}

\scriptsize
\setlength{\parskip}{0pt}

\begin{tcolorbox}[colback=green!5, colframe=green!40, halign=flush left, title={\scriptsize\textbf{FoodSense-VL (Stage~2)} --- Taste:4.0\, Smell:3.5\, Texture:3.0\, Sound:2.5}, left=1pt, right=1pt, top=0.5pt, bottom=0.5pt, boxrule=0.3pt]
\textbf{Taste:} The deep, dark color of the steak suggests a \underline{robust}, \underline{savory} flavor with a slightly \underline{charred} tone. The red rice and visible garnish imply a \underline{spicy}, tomato-forward base. Onion rings add \underline{pungency} and contrast, and the plated components suggest a \underline{bold}, \underline{well-seasoned} profile.\\
\textbf{Smell:} Grill marks and sheen on the steak suggest a \underline{sizzling}, \underline{savory aroma}. The rice color cues \underline{warm spice} notes, while lime and green onion suggest a \underline{bright citrus} lift. Overall smell appears \underline{rich} and layered.\\
\textbf{Texture:} The steak looks \underline{firm} and \underline{chewy}, with \underline{fluffy} separated rice grains. The tortilla/chip element appears \underline{brittle} and \underline{crunchy}, while onion rings add slight bite resistance.\\
\textbf{Sound:} The dish likely stays relatively \underline{quiet} overall, with most sound from a \underline{crisp snap} in the tortilla/chip element and subtle \underline{rustling} from rice.
\end{tcolorbox}

\vspace{0pt}
\begin{tcolorbox}[colback=purple!5, colframe=purple!50, halign=flush left, title={\scriptsize\textbf{Qwen2.5-VL~32B} --- Taste:4.5\, Smell:4.0\, Texture:4.0\, Sound:2.5}, left=1pt, right=1pt, top=0.5pt, bottom=0.5pt, boxrule=0.3pt]
\textbf{Taste:} The dish appears \underline{rich} and \underline{flavorful}, with \underline{savory}, slightly \underline{smoky} grilled meat, \underline{warm earthy} orange rice, and a \underline{creamy} side component. Pickled onions add a \underline{tangy}, \underline{sharp} contrast.\\
\textbf{Smell:} The profile is likely \underline{robust}, led by the \underline{charred aroma} of grilled meat and \underline{fragrant spices}. Lime and fresh vegetables suggest a \underline{citrusy} lift, while pickled onions add a \underline{pungent vinegary} note.\\
\textbf{Texture:} Expected texture is \underline{varied}: \underline{tender-juicy} meat with a slightly \underline{crispy charred} exterior, \underline{fluffy moist} rice, a \underline{creamy-smooth} side, and \underline{crisp crunchy} onion accents.\\
\textbf{Sound:} Auditory cues are mostly \underline{subtle}, with imagined \underline{sizzle} from grilled meat, occasional \underline{crunch} from onions, and \underline{soft chew} from rice and side components.
\end{tcolorbox}

\begin{tcolorbox}[colback=brown!5, colframe=brown!50, halign=flush left, title={\scriptsize\textbf{Food-Llama~11B} --- Taste:4.0\, Smell:4.0\, Texture:4.0\, Sound:3.0}, left=1pt, right=1pt, top=0.5pt, bottom=0.5pt, boxrule=0.3pt]
\textbf{Taste:} The plate suggests a \underline{rich} \underline{savory} profile from \underline{grilled} steak and \underline{seasoned} rice, with \underline{tang} from onion/lime garnish and \underline{creamy} contrast from cheese-covered tortilla.\\
\textbf{Smell:} Likely \underline{smoky} and \underline{warm} from the \underline{grilled} meat, with a \underline{spiced} rice \underline{aroma} and \underline{fresh citrus-herb} top notes.\\
\textbf{Texture:} Expected texture is mixed: \underline{tender-chewy} steak, \underline{fluffy} rice, \underline{crunchy} tortilla/chip components, and \underline{creamy-soft} topping elements.\\
\textbf{Sound:} Audible cues are \underline{moderate}, mainly from \underline{crisp tortilla crunch} and light utensil sounds, with \underline{softer chew} from meat and rice.
\end{tcolorbox}

\vspace{0pt}
\begin{tcolorbox}[halign=flush left, colback=red!5, colframe=red!50, title={\scriptsize\textbf{LLaVA} --- Taste:3.5\, Smell:4.0\, Texture:3.0\, Sound:3.0}, left=2pt, right=2pt, top=1pt, bottom=1pt, boxrule=0.3pt]
\textbf{Taste:} The dish appears Mexican-inspired, with \underline{savory} meat, sauce, and rice as the core flavor base. Onion and tomato elements suggest \underline{fresh tang} and contrast.\\
\textbf{Smell:} Likely \underline{aromatic} and \underline{savory}, with grilled-meat notes and \underline{spiced} sauce. Rice adds a \underline{mild earthy} background smell.\\
\textbf{Texture:} Expected texture mix is \underline{tender} meat, \underline{soft} rice, and some \underline{chew} from tortilla components, with occasional \underline{crispness} from garnish.\\
\textbf{Sound:} Predominantly \underline{gentle} eating sounds, with occasional \underline{crunch} from tortilla elements and \underline{soft} utensil contact.
\end{tcolorbox}
\vspace{0pt}
\vspace{-6pt}
\caption{Qualitative sensory inferences for \textbf{Steak Rice} from four models. Human GT: Taste=4.3, Smell=4.3, Texture=4.4, Sound=4.1.}
\label{fig:qual_comparison}
\vspace{-0.35cm}
\end{figure}

\subsection{Quantitative Insights}
\label{sec:quant_results}

\noindent\textbf{Insight 1: Low Absolute Error Can Mask Poor Discrimination.} 
While generalist VLMs like InternVL-2.5 achieve the lowest absolute errors (MAE/RMSE) on modalities like Sound, this is a symptom of outputting safe, average ratings heavily clustered around 3.5--4.0. This behavior is common in large language models evaluated on out-of-distribution subjective tasks, where models retreat to mean values to minimize penalty when lacking genuine discriminative understanding. In contrast, our fine-tuned model actively predicts extreme values (e.g., $1.0$ for quiet foods, $4.5$ for loud foods), which incurs a higher average absolute penalty but correctly models human rating variance.

\noindent\textbf{Insight 2: Correlation and CCC Reveal True Cross-Sensory Understanding.}
Because absolute errors easily reward ``averaging'' behavior, correlation-based metrics provide a more informative assessment. FoodSense-VL achieves the highest overall Pearson~$r = 0.372$, Spearman~$\rho = 0.360$, and Lin's CCC $= 0.343$ (Tables~\ref{tab:sensory_rank}--\ref{tab:sensory_pearson}), with CCC exceeding the next-best baseline (Gemma 3 base, 0.136) by over 150\%. Lin's CCC is particularly revealing because it jointly penalizes both poor correlation and scale bias.

\noindent\textbf{Insight 3: Two-Stage Curriculum Trades Conservatism for Discrimination.}
Table~\ref{tab:ablation} ablates the two-stage design by comparing against a single-stage model that sees the same combined data (human $+$ MAmmoTH expansion) but trains from a fresh LoRA in one pass. The two-stage curriculum improves Pearson~$r$ by $+0.043$, Spearman~$\rho$ by $+0.047$, and CCC by $+0.069$, while increasing prediction diversity ($\sigma_{\text{pred}}$: $0.367 \to 0.591$). The MAE increases by only $+0.044$---a trade-off we consider affordable. Separating sensory grounding (Stage~1) from reasoning integration (Stage~2) encourages the model to spread its predictions across the rating scale rather than hedge toward the mean.

\subsection{Qualitative Observations}
\label{sec:qual_results}
\vspace{-0.15cm}
\noindent\textbf{Observation 1: Sound is the Hardest Modality to Infer.} 
Inferring the auditory experience of biting into food from a static 2D image is notoriously difficult. Our human annotations show high variance in sound ratings. We observe substantial cross-model variation on this dimension: some models are overly conservative and collapse toward mid-scale predictions, while others overestimate audible texture cues. This aligns with findings that current multimodal architectures can over-rely on prominent visual features and struggle with non-salient, cross-modal reasoning~\cite{GUNNING2025101127}. In our benchmark, sound remains a key differentiator across open-source baselines and FoodSense-VL, and qualitative outputs show that explicit sensory grounding improves when models tie sound judgments to visible structural cues (e.g., crispy edges, brittle coatings, and layered textures).

\noindent\textbf{Observation 2: Descriptive Richness Translates to Interpretability.}
Fig.~\ref{fig:qual_comparison} compares Steak Rice inferences from FoodSense-VL, Qwen2.5-VL, Food-LLaMA, and LLaVA. A key advantage of training on our sensory dataset is the generation of food-specific sensory vocabulary (underlined) grounded in explicit visual cues. FoodSense-VL and Qwen provide richer, cue-linked sensory justifications, while LLaVA and Food-LLaMA remain comparatively generic and less discriminative in cross-sensory grounding.

\section{Conclusion}
\label{sec:conclusion}
\vspace{-0.3cm}
We presented a dataset and pipeline for predicting taste, smell, texture, 
and sound from food images. Human annotations are expanded into reasoning traces, and two-stage fine-tuning yields both ratings and explanations. Our model achieves the best Pearson~$r$ ($0.372$), Spearman~$\rho$ ($0.360$), and Lin's CCC ($0.343$) among all evaluated VLMs, demonstrating genuine discriminative understanding of cross-sensory properties. Our ablation study shows that two-stage training improves CCC by $+0.069$ over matched single-stage training. Sound remains one of the most challenging modality for all models evaluated, consistent with its lower ground-truth inter-rater reliability. Future work will evaluate whether generated rationales align with human judgments and examine cultural moderators of cross-sensory inference.
\FloatBarrier
{\small
\bibliographystyle{ieeenat_fullname}
\bibliography{main}
}


\clearpage
\appendix
\setcounter{table}{0}
\renewcommand{\thetable}{\Alph{section}\arabic{table}}
\setcounter{figure}{0}
\renewcommand{\thefigure}{\Alph{section}\arabic{figure}}


\section{FoodSense Annotation Protocol}
\label{sec:suppl_annotation}

\subsection{Task Design}
\label{sec:suppl_task}

Participants were shown one food image at a time and asked to evaluate four sensory dimensions: \textbf{taste}, \textbf{smell}, \textbf{texture}, and \textbf{sound}. For each dimension, participants completed two sub-tasks sequentially.

\noindent\textbf{Quantitative rating.} The survey prompt read: \textit{``Based on the image above, how would you rate the likely [taste / smell / texture / sound] of this food?''} Responses were recorded on the seven-point scale shown in Table~\ref{tab:supp_likert}. Participants could also select a \textit{Can't tell from picture} option (coded as 0) whenever the image provided insufficient visual cues.

\noindent\textbf{Qualitative descriptor.} After rating, participants were asked: \textit{``What do you think this food would sound like, taste like, smell like, and feel like (texture)? Please write \underline{one or two words} for each sense.''} Representative responses include \textit{crispy}, \textit{golden edges}, \textit{smoky}, and \textit{silent}.

This dual-format design captures both the magnitude of anticipated sensory experience and the natural language people use to ground those judgments in visual evidence. Valid numeric ratings were linearly rescaled from the original 1--7 range to a 1--5 scale via the transformation $r_k = 1 + [(r_{\mathrm{orig}} - 1) \times 4] / 6$, preserving relative ordering. Responses marked as \textit{Can't tell from picture} were excluded from rescaling and retained as a separate binary \texttt{CanInfer}$_k$ flag per dimension.

\begin{table}[h]
\centering
\caption{Seven-point Likert scale used for sensory ratings.}
\label{tab:supp_likert}
\begin{tabular}{@{}cl@{}}
\toprule
\textbf{Value} & \textbf{Meaning} \\
\midrule
0 & Can't tell from picture \\
1 & Very bad \\
2 & Bad \\
3 & Slightly bad \\
4 & Neutral \\
5 & Slightly good \\
6 & Good \\
7 & Very good \\
\bottomrule
\end{tabular}
\end{table}

\subsection{Survey Interface}
\label{sec:suppl_interface}

\begin{figure}[h]
    \centering
    \includegraphics[width=0.75\linewidth]{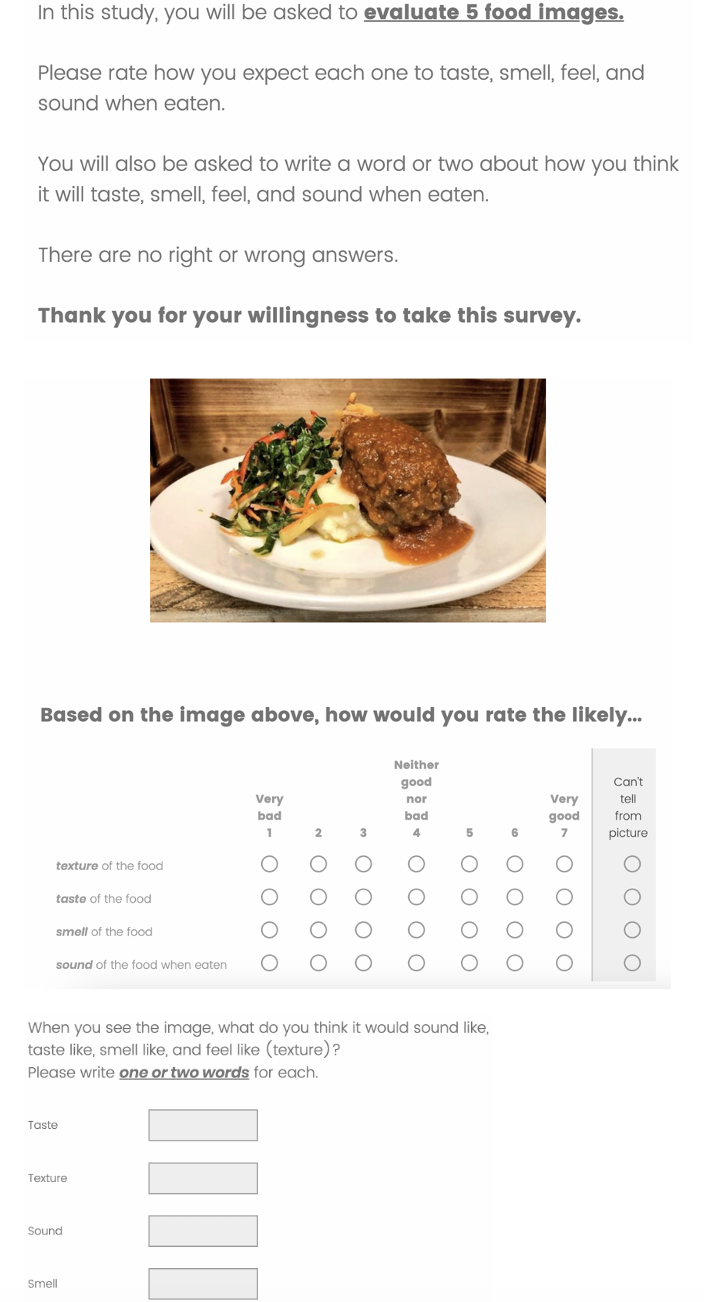}
    \caption{Qualtrics survey interface as presented to participants, showing the food image display, seven-point rating scales for each sensory dimension, and free-text descriptor entry fields.}
    \label{fig:supp_survey}
\end{figure}

The annotation survey was administered via Qualtrics. Each survey page presented a single food image at the top, followed by four rating scale questions (one per sensory dimension) and four free-text response fields. Figure~\ref{fig:supp_survey} shows a representative screenshot of the survey interface as presented to participants.

\subsection{Participant Recruitment}
\label{sec:suppl_recruitment}

Participants were recruited through two channels: (i) a professional online survey panel ($n = 7{,}734$; 63,741 annotations, 95.4\%) and (ii) a university behavioral laboratory ($n = 648$; 3,101 annotations, 4.6\%). In total, 8,382 participants contributed 66,842 assessments across 2,987 images (mean $= 22.38$ annotations per image, $SD = 2.02$). The dual-channel recruitment strategy strengthens annotation quality---laboratory participants provide controlled, distraction-free responses, while the large online panel ensures scale and demographic diversity. No demographic information was collected as part of the annotation protocol.

Following quality filtering---removing 72 images due to filename inconsistencies in the training pipeline and retaining only assessments where participants indicated visual inferability for all four dimensions (\texttt{CanInfer}$_k = 1$)---the final training dataset comprises 58,443 annotations across 2,915 images.


\section{Additional FoodSense Dataset Statistics}
\label{sec:suppl_stats}

\subsection{Annotator Distribution}
\label{sec:suppl_annotator_dist}

\begin{figure}[h]
    \centering
    \includegraphics[width=0.75\linewidth]{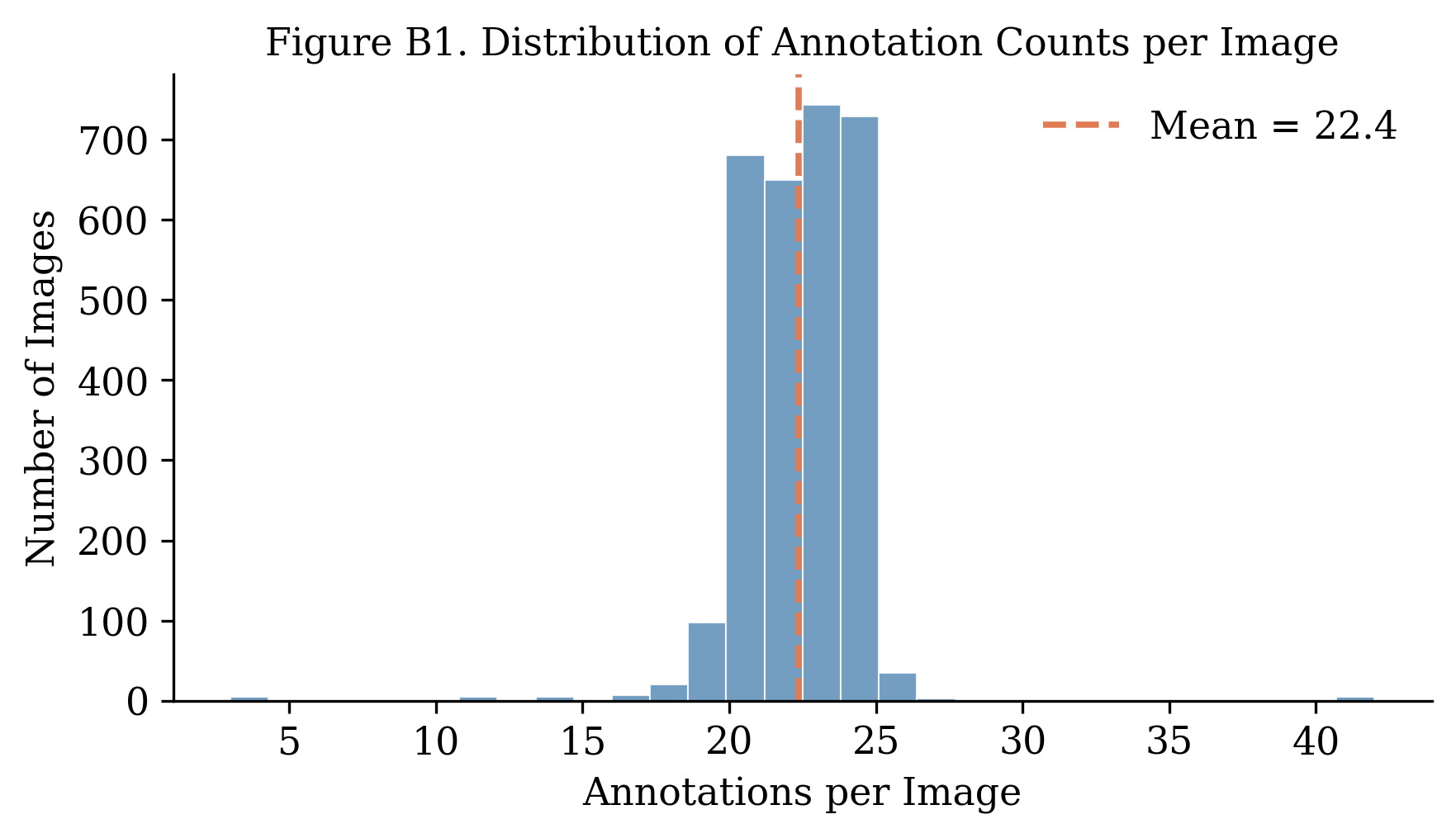}
    \caption{Distribution of annotation counts per image across the full dataset of 2,987 images. The dashed line indicates the mean (22.38 annotations per image).}
    \label{fig:supp_annotator_hist}
\end{figure}

Table~\ref{tab:supp_annotator} summarizes the distribution of annotation 
counts across the full released dataset of 2,987 images. The distribution 
is approximately Gaussian, centered near 22 annotations per image 
(Figure~\ref{fig:supp_annotator_hist}), with a long right tail attributable 
to variability in the number of participants assigned to each image across 
survey sets.

\begin{table}[h]
\centering
\caption{Summary statistics for annotation counts across 2,987 images.}
\label{tab:supp_annotator}
\begin{tabular}{@{}lc@{}}
\toprule
\textbf{Statistic} & \textbf{Value} \\
\midrule
Mean annotations per image       & 22.38 \\
Standard deviation               & 2.02  \\
Minimum                          & 3     \\
Maximum                          & 42    \\
Total annotations (full dataset) & 66,842 \\
Total annotations (training set) & 58,443 \\
\bottomrule
\end{tabular}
\end{table}

\subsection{Data Partitioning}
\label{sec:suppl_splits}

To construct train/validation/test splits, we applied pseudo-random stratified shuffle splitting at the image level (75\% / 10\% / 15\%) based on binned mean overall rating, ensuring all three partitions share equivalent sensory rating distributions. All splits are strictly image-level---no image appears in more than one partition. Table~\ref{tab:supp_splits} summarizes the final partition sizes.

\begin{table}[h]
\centering
\caption{Dataset partitions with image and annotation counts.}
\label{tab:supp_splits}
\begin{tabular}{@{}lrrr@{}}
\toprule
\textbf{Split} & \textbf{Images} & \textbf{Annotations} & \textbf{\% of Total} \\
\midrule
Train      & 2,185 & 43,758 & 74.9\% \\
Validation & 292   & 5,834  & 10.0\% \\
Test       & 438   & 8,851  & 15.2\% \\
\midrule
Total      & 2,915 & 58,443 & 100\%  \\
\bottomrule
\end{tabular}
\end{table}

\subsection{Descriptor Vocabulary}
\label{sec:suppl_vocab}

\begin{figure}[h]
    \centering
    \includegraphics[width=0.85\linewidth]{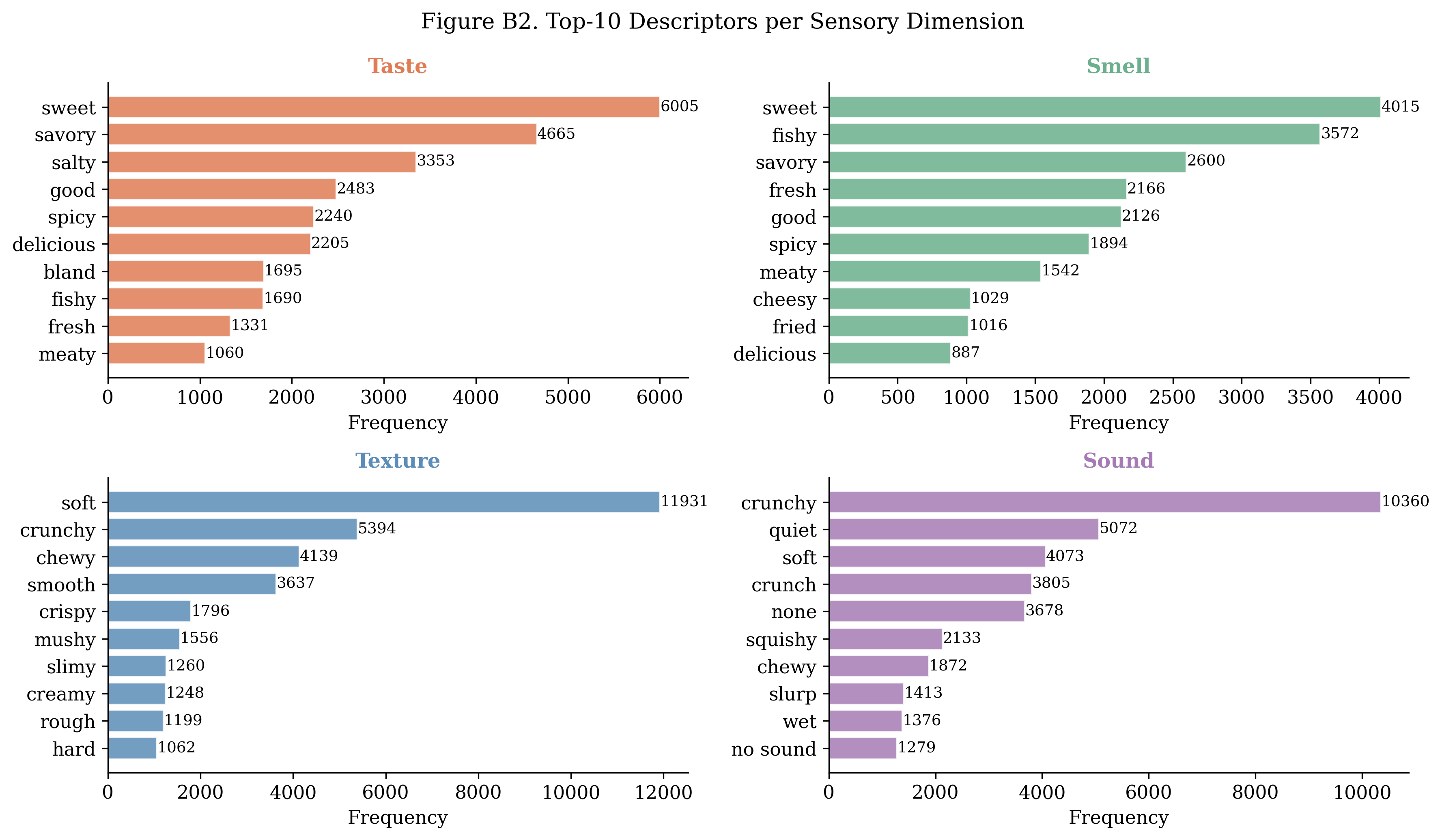}
    \caption{Top-10 descriptor frequencies per sensory dimension across all 2,987 images.}
    \label{fig:supp_descriptors}
\end{figure}

Participants provided one to two free-text words per sensory dimension. The descriptor vocabulary spans 25,508 unique terms across 265,915 total entries (including repetitions across participants). Table~\ref{tab:supp_vocab} reports unique vocabulary sizes and representative high-frequency terms per dimension. Figure~\ref{fig:supp_descriptors} shows the top-10 descriptor frequencies for each sense.

\begin{table}[h]
\centering
\caption{Descriptor vocabulary statistics and top-10 terms by sensory dimension.}
\label{tab:supp_vocab}
\resizebox{\columnwidth}{!}{%
\begin{tabular}{@{}llp{8cm}@{}}
\toprule
\textbf{Dimension} & \textbf{Unique Terms} & \textbf{Top 10 Descriptors} \\
\midrule
Taste   & 9,217 & sweet, savory, salty, good, spicy, delicious, bland, fishy, fresh, meaty \\
Smell   & 9,161 & sweet, fishy, savory, fresh, good, spicy, meaty, cheesy, fried, delicious \\
Texture & 6,807 & soft, crunchy, chewy, smooth, crispy, mushy, slimy, creamy, rough, hard \\
Sound   & 5,323 & crunchy, quiet, soft, crunch, none, squishy, chewy, slurp, wet, no sound \\
\bottomrule
\end{tabular}%
}
\end{table}

\subsection{Cross-Sensory Correlations}
\label{sec:suppl_correlations}

Table~\ref{tab:supp_corr} reports Pearson correlations between per-image mean ratings across all four sensory dimensions ($N = 2{,}987$ images, \texttt{CanInfer}$_k = 1$ only, all $p < .001$). Correlations are uniformly strong and positive, consistent with a shared visual appetitiveness signal driving ratings across dimensions. The weakest pairings involve sound ($r = 0.726$--$0.840$), reflecting the greater difficulty of auditory inference from static images---a pattern consistent with the lower ICC(1,k) for sound reported in the main paper.

\begin{table}[h]
\centering
\caption{Pearson correlations between per-image mean sensory ratings ($N = 2{,}987$; all $p < .001$).}
\label{tab:supp_corr}
\begin{tabular}{@{}lcccc@{}}
\toprule
              & \textbf{Taste} & \textbf{Smell} & \textbf{Texture} & \textbf{Sound} \\
\midrule
\textbf{Taste}   & ---   & 0.893 & 0.931 & 0.759 \\
\textbf{Smell}   & 0.893 & ---   & 0.856 & 0.726 \\
\textbf{Texture} & 0.931 & 0.856 & ---   & 0.840 \\
\textbf{Sound}   & 0.759 & 0.726 & 0.840 & ---   \\
\bottomrule
\end{tabular}
\end{table}

\begin{figure}[h]
    \centering
    \includegraphics[width=0.55\linewidth]{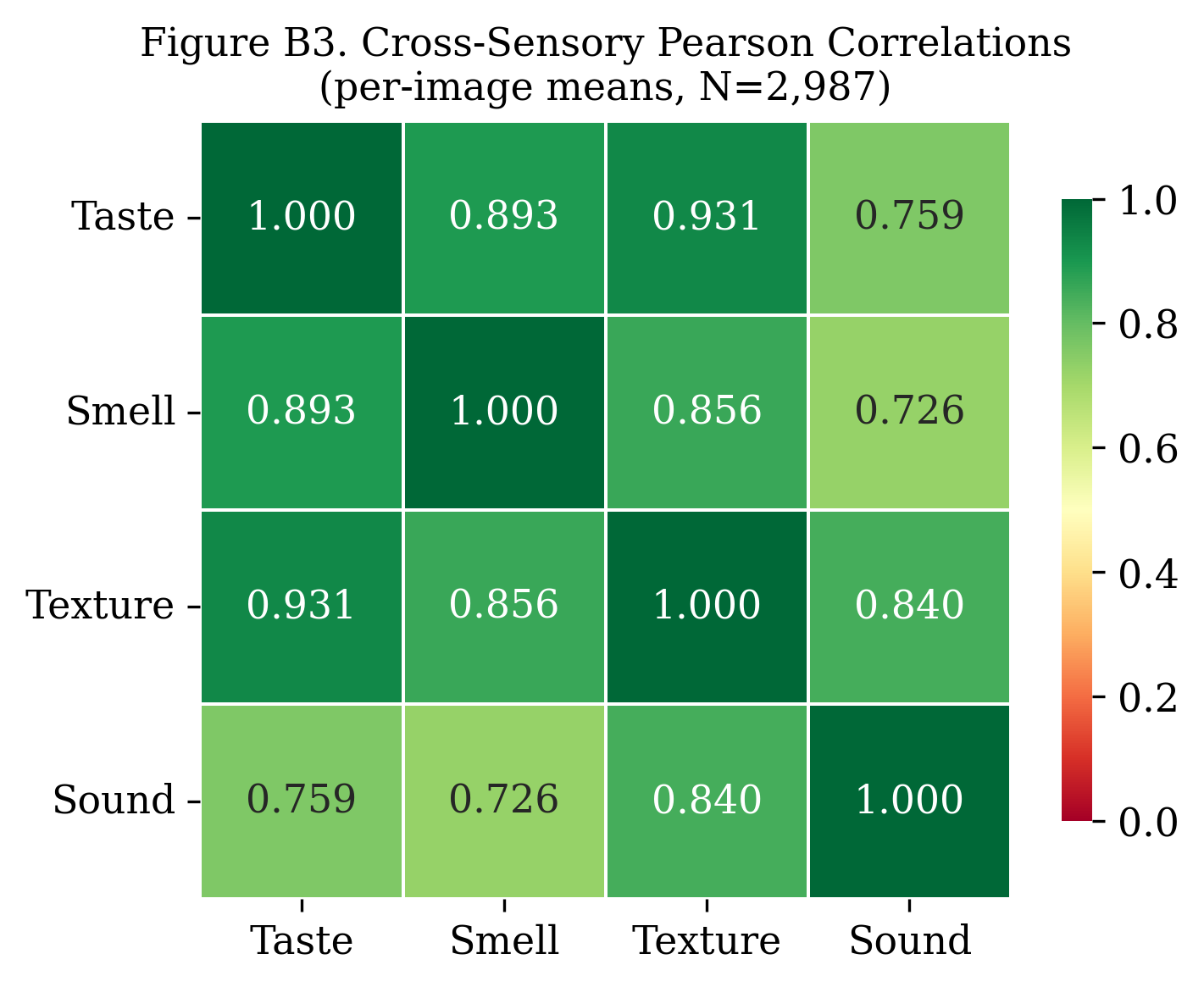}
    \caption{Heatmap of Pearson correlations between per-image mean sensory ratings ($N = 2{,}987$).}
    \label{fig:supp_heatmap}
\end{figure}

\subsection{Descriptor Consistency and Uncertainty Structure}
\label{sec:suppl_desc_consistency}

To assess annotation consistency in the free-text descriptors, we computed per-image lexical diversity and coverage statistics for each sensory dimension, excluding missing responses and uncertainty expressions (e.g., ``not sure'', ``can't tell''). Table~\ref{tab:supp_desc_diversity} reports the mean type-token ratio (TTR; unique descriptors divided by total valid descriptors per image), top-1 coverage (fraction of annotations using the most common descriptor), and top-3 coverage per sense.

\begin{table}[h]
\centering
\caption{Per-image descriptor diversity and coverage statistics across 2,987 images. TTR = type-token ratio (higher = more diverse). Top-$k$ coverage = fraction of annotations using one of the $k$ most common descriptors for that image.}
\label{tab:supp_desc_diversity}
\resizebox{\columnwidth}{!}{%
\begin{tabular}{@{}lcccc@{}}
\toprule
\textbf{Dimension} & \textbf{Mean Unique} & \textbf{Mean TTR} & \textbf{Top-1 Coverage} & \textbf{Top-3 Coverage} \\
\midrule
Taste   & 16.04 (3.22) & 0.725 (0.137) & 0.218 & 0.396 \\
Smell   & 16.75 (3.26) & 0.760 (0.135) & 0.200 & 0.367 \\
Texture & 14.60 (2.84) & 0.658 (0.120) & 0.245 & 0.457 \\
Sound   & 14.43 (3.08) & 0.663 (0.134) & 0.232 & 0.446 \\
\bottomrule
\end{tabular}%
}
\end{table}

TTR values ranging from 0.658 to 0.760 indicate substantial lexical diversity across all dimensions, consistent with the inherently subjective nature of sensory inference from images. Texture and sound show slightly lower TTR and higher top-3 coverage than taste and smell, suggesting modestly greater convergence on a smaller set of perceptual anchors (e.g., \textit{soft}, \textit{crunchy} for texture; \textit{quiet}, \textit{crunchy} for sound). However, top-3 coverage does not exceed 0.457 for any dimension, confirming that no small cluster of terms dominates annotations for a given image.

We further examined whether descriptor diversity and numeric rating disagreement capture the same underlying uncertainty signal. Table~\ref{tab:supp_ttr_sd} reports Pearson correlations between per-image TTR and per-image rating SD across all four dimensions.

\begin{table}[h]
\centering
\caption{Pearson correlations between per-image descriptor TTR and rating SD ($N = 2{,}987$). Higher TTR indicates more diverse descriptors; higher rating SD indicates more numeric disagreement.}
\label{tab:supp_ttr_sd}
\begin{tabular}{@{}lcc@{}}
\toprule
\textbf{Dimension} & \textbf{Pearson $r$} & \textbf{$p$-value} \\
\midrule
Taste   & 0.249 & $<.0001$ \\
Smell   & $-$0.047 & $.011$ \\
Texture & 0.141 & $<.0001$ \\
Sound   & 0.068 & $.0002$ \\
\bottomrule
\end{tabular}
\end{table}

For taste and texture, descriptor diversity and rating variance covary positively and significantly, indicating that the two annotation modalities capture a shared uncertainty signal: images that elicit more varied numeric ratings also attract more diverse textual descriptions. Sound and smell show weaker effects. For sound specifically, the \texttt{CanInfer} rate is also uncorrelated with rating SD ($r = 0.007$, $p = .71$), unlike the other three dimensions where higher inferability rates associate with lower rating variance (taste: $r = -0.160$; smell: $r = -0.137$; texture: $r = -0.112$; all $p < .0001$). Together, these patterns suggest that auditory inference from static images operates through a qualitatively different uncertainty mechanism than the other sensory dimensions, consistent with sound's lower ICC(1,k) reported in the main paper.


\section{Additional FoodSense Annotation Examples}
\label{sec:suppl_examples}

This section presents two representative annotation examples drawn from the test set (sampled randomly, seed~$= 42$). Each entry reports the per-image mean rescaled rating (1--5 scale) and the primary free-text descriptor for each sensory dimension.

\begin{figure}[H]
    \centering
    \includegraphics[width=\linewidth]{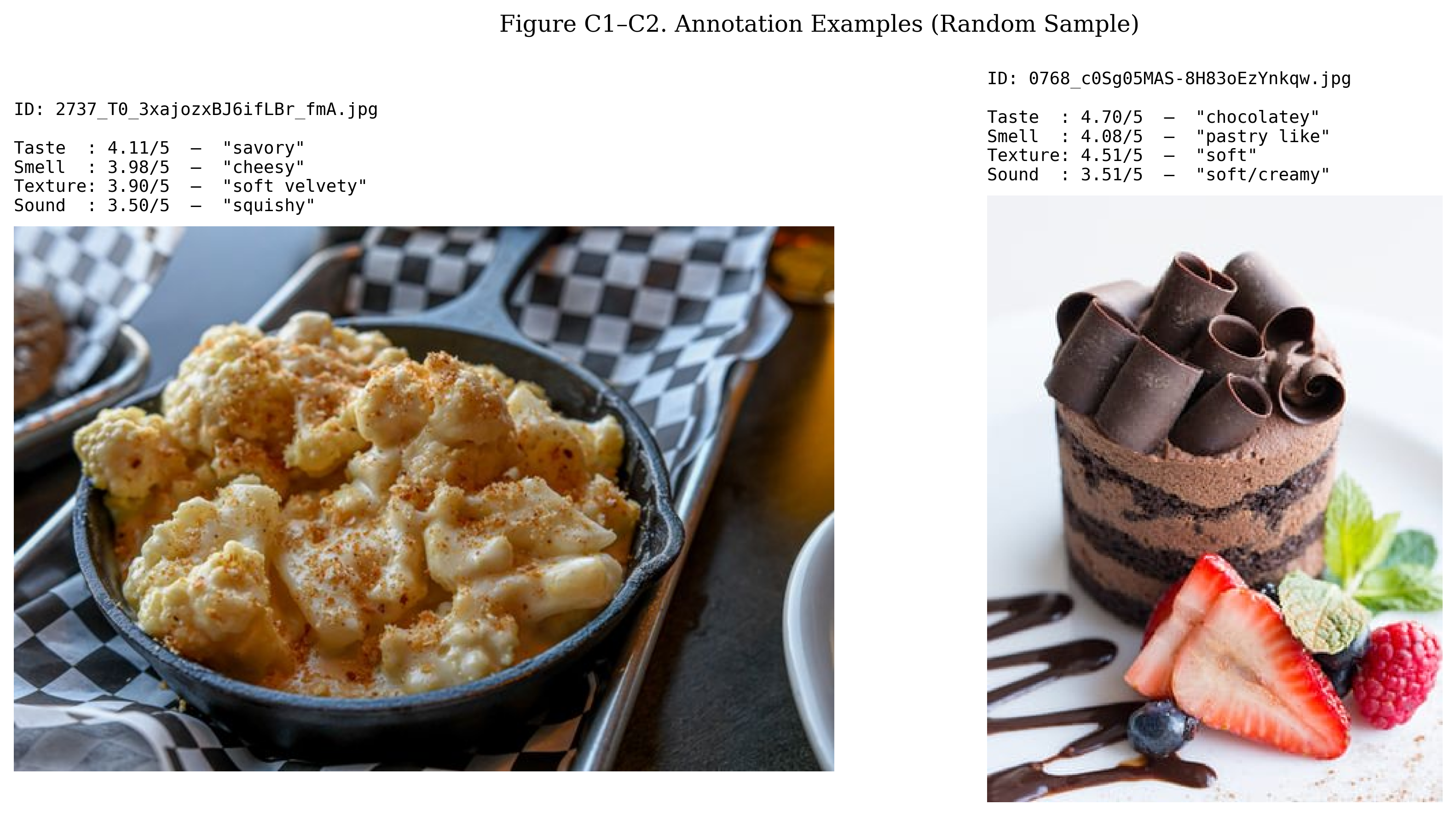}
    \caption{Two annotation examples randomly sampled from the test set (seed = 42). Left: Image 2737 (mac and cheese). Right: Image 0768 (chocolate dessert).}
    \label{fig:supp_examples}
\end{figure}

\noindent\textbf{Example 1 --- Image 2737.} Taste: 4.11 (\textit{savory}); Smell: 3.98 (\textit{cheesy}); Texture: 3.90 (\textit{soft velvety}); Sound: 3.50 (\textit{squishy}).

\noindent\textbf{Example 2 --- Image 0768.} Taste: 4.70 (\textit{chocolatey}); Smell: 4.08 (\textit{pastry like}); Texture: 4.51 (\textit{soft}); Sound: 3.51 (\textit{soft/creamy}).


\section{Extended Ablation \& Rating Distribution Analysis}
\label{sec:suppl_distribution}

This appendix provides additional evidence for the claims in Sec.~5 of the main paper, examining how per-image rating distributions differ across models and how the two-stage curriculum reshapes prediction behavior.

\subsection{Prediction Spread ($\sigma_{\text{pred}}$)}
\label{sec:suppl_sigma}

Table~\ref{tab:suppl_sigma} reports the standard deviation of each model's predicted ratings, broken down by sense. A model that collapses toward a constant value will have $\sigma_{\text{pred}} \approx 0$; a model that utilizes the full 1--5 scale will approach or exceed the human ground-truth spread.

\begin{table}[h]
\centering
\caption{Prediction standard deviation ($\sigma_{\text{pred}}$) per sense and overall. Human GT row shows the spread of ground-truth mean ratings across the 438 test images. \textbf{Bold} = closest to GT spread.}
\label{tab:suppl_sigma}
\renewcommand{\arraystretch}{1.2}
\resizebox{\columnwidth}{!}{%
\begin{tabular}{@{}lccccc@{}}
\toprule
Model & Taste & Smell & Texture & Sound \\
\midrule
Human GT                    & 0.502 & 0.518 & 0.498 & 0.385 \\
\midrule
FoodSense-VL (Two-Stage)   & \textbf{0.499} & \textbf{0.488} & 0.759 & 0.619 \\
Single-Stage (lr=$2{\times}10^{-6}$) & 0.236 & 0.392 & \textbf{0.490} & 0.351 \\
Gemma~3 27B (base)          & 0.252 & 0.382 & 0.461 & 0.3498 \\
LLaVA-v1.6-34B              & 0.155 & 0.252 & 0.149 & \textbf{0.398} \\
InternVL2.5-26B             & 0.083 & 0.075 & 0.245 & 0.181 \\
Qwen2.5-VL-32B              & 0.278 & 0.253 & 0.293 & 0.560 \\
Food-LLaMA-11B              & 0.539 & 0.750 & 0.421 & 0.889 \\
\bottomrule
\end{tabular}%
}
\end{table}

FoodSense-VL's taste $\sigma_{\text{pred}} = 0.499$ nearly matches the human GT spread of $0.502$, indicating that the two-stage curriculum successfully learns to use the full rating scale. By contrast, InternVL collapses to $\sigma_{\text{pred}} = 0.083$ for taste and $0.075$ for smell---effectively predicting a near-constant value. The single-stage ablation shows intermediate spread ($0.236$ for taste), confirming that the staged curriculum is responsible for the increased prediction diversity rather than the training data alone.

\subsection{Rating Bin Distribution}
\label{sec:suppl_bins}

Table~\ref{tab:suppl_bins} shows the percentage of predictions falling into each integer-centered bin ($\pm 0.5$), along with Shannon entropy as a diversity measure. Higher entropy indicates more uniform use of the rating scale.

\begin{table}[h]
\centering
\caption{Distribution of predicted ratings across bins (percentage of all predictions). Entropy computed as $H = -\sum p_i \log_2 p_i$ over the 5 bins.}
\label{tab:suppl_bins}
\renewcommand{\arraystretch}{1.2}
\resizebox{\columnwidth}{!}{%
\begin{tabular}{@{}lcccccc@{}}
\toprule
Model & ${\sim}1$ & ${\sim}2$ & ${\sim}3$ & ${\sim}4$ & ${\sim}5$ & Entropy \\
\midrule
Human GT                    & 0.0\% & 1.5\% & 30.2\% & 63.5\% & 4.7\%  & 1.239 \\
\midrule
FoodSense-VL (Two-Stage)   & 1.0\% & 8.9\% & 29.7\% & 48.5\% & 12.0\% & 1.769 \\
Single-Stage (lr=$2{\times}10^{-6}$) & 0.1\% & 3.5\% & 30.6\% & 62.9\% & 3.0\% & 1.269 \\
Gemma~3 27B (base)          & 0.1\% & 7.0\% & 26.3\% & 63.9\% & 2.7\%  & 1.337 \\
LLaVA-v1.6-34B              & 0.0\% & 0.6\% & 11.1\% & 63.9\% & 24.4\% & 1.307 \\
InternVL2.5-26B             & 0.0\% & 0.2\% & 2.0\%  & 64.3\% & 33.6\% & 1.066 \\
Qwen2.5-VL-32B              & 0.9\% & 1.0\% & 13.7\% & 53.1\% & 31.4\% & 1.526 \\
Food-LLaMA-11B              & 2.8\% & 13.4\% & 20.4\% & 44.9\% & 18.4\% & 1.970 \\
\bottomrule
\end{tabular}%
}
\end{table}

FoodSense-VL has the highest entropy (1.77) among models that also achieve strong correlation, indicating it balances prediction diversity with discrimination. InternVL has the lowest entropy (1.07), concentrating 97.9\% of its predictions in the 4--5 range, which explains its low MAE (close to the mean) but poor correlation.

\subsection{Per-Sense Mean Ratings}
\label{sec:suppl_means}

Table~\ref{tab:suppl_means} compares per-sense mean predictions across models. Systematic deviations from human GT means indicate calibration bias.

\begin{table}[h]
\centering
\caption{Mean predicted rating $\pm$ std per sensory dimension. Human GT shows the mean of annotator-averaged ratings.}
\label{tab:suppl_means}
\renewcommand{\arraystretch}{1.2}
\resizebox{\columnwidth}{!}{%
\begin{tabular}{@{}lcccc@{}}
\toprule
Model & Taste & Smell & Texture & Sound \\
\midrule
Human GT                    & $3.88 \pm 0.50$ & $3.80 \pm 0.52$ & $3.74 \pm 0.50$ & $3.44 \pm 0.39$ \\
\midrule
FoodSense-VL (Two-Stage)   & $3.88 \pm 0.50$ & $3.48 \pm 0.49$ & $3.64 \pm 0.76$ & $2.62 \pm 0.62$ \\
Single-Stage                & $3.82 \pm 0.24$ & $3.57 \pm 0.39$ & $3.81 \pm 0.49$ & $2.51 \pm 0.35$ \\
Gemma~3 27B (base)          & $3.85 \pm 0.25$ & $3.56 \pm 0.38$ & $3.85 \pm 0.46$ & $2.38 \pm 0.35$ \\
LLaVA-v1.6-34B              & $4.47 \pm 0.16$ & $3.90 \pm 0.25$ & $3.98 \pm 0.15$ & $3.20 \pm 0.40$ \\
InternVL2.5-26B             & $4.49 \pm 0.08$ & $4.00 \pm 0.07$ & $4.17 \pm 0.25$ & $3.45 \pm 0.18$ \\
Qwen2.5-VL-32B              & $4.29 \pm 0.28$ & $4.00 \pm 0.25$ & $4.26 \pm 0.29$ & $3.06 \pm 0.56$ \\
Food-LLaMA-11B              & $3.82 \pm 0.54$ & $3.44 \pm 0.75$ & $4.09 \pm 0.42$ & $2.50 \pm 0.89$ \\
\bottomrule
\end{tabular}%
}
\end{table}

FoodSense-VL's taste mean ($3.88$) matches the human GT exactly, while generalist VLMs systematically over-predict (InternVL: $4.49$, LLaVA: $4.47$). Sound is consistently under-predicted by all fine-tuned models (FoodSense-VL: $2.62$ vs.\ GT: $3.44$), reflecting the inherent difficulty of auditory inference from static images.

\subsection{Kolmogorov--Smirnov Distribution Tests}
\label{sec:suppl_ks}

Table~\ref{tab:suppl_ks} reports pairwise two-sample Kolmogorov--Smirnov test statistics ($D$) between each model pair. Larger $D$ indicates more divergent rating distributions. All pairs marked with $^*$ are significant at $p < 0.05$.

\begin{table}[h]
\centering
\caption{Pairwise KS $D$-statistics across all model rating distributions (all senses pooled). $^*$\,$p < 0.05$.}
\label{tab:suppl_ks}
\renewcommand{\arraystretch}{1.1}
\resizebox{\columnwidth}{!}{%
\begin{tabular}{@{}lccccccc@{}}
\toprule
 & Ours & Single & Base & LLaVA & IntVL & Qwen & F-LLaMA \\
\midrule
Ours (Two-Stage)  & ---    & .148$^*$ & .134$^*$ & .312$^*$ & .374$^*$ & .338$^*$ & .125$^*$ \\
Single-Stage      & ---    & ---      & .036     & .460$^*$ & .517$^*$ & .487$^*$ & .274$^*$ \\
Base              & ---    & ---      & ---      & .446$^*$ & .502$^*$ & .472$^*$ & .259$^*$ \\
LLaVA             & ---    & ---      & ---      & ---      & .096$^*$ & .071$^*$ & .249$^*$ \\
InternVL          & ---    & ---      & ---      & ---      & ---      & .133$^*$ & .345$^*$ \\
Qwen              & ---    & ---      & ---      & ---      & ---      & ---      & .213$^*$ \\
\bottomrule
\end{tabular}%
}
\end{table}

Key observations: (1)~The single-stage ablation and base model have nearly identical distributions ($D = 0.036$, not significant), suggesting that without the staged curriculum, the model does not meaningfully reshape its output distribution. (2)~FoodSense-VL's distribution is significantly different from both the single-stage ($D = 0.148$) and the base ($D = 0.134$), confirming the curriculum effect. (3)~Generalist VLMs (LLaVA, InternVL, Qwen) cluster together with small pairwise distances ($D \leq 0.133$) but diverge sharply from fine-tuned models ($D > 0.3$).



\section{Full Sensory Inference Comparison}
\label{sec:suppl}

This supplementary material provides the full sensory ratings table and complete text inferences from all models across six representative food images. Sensory keywords are \underline{underlined}. Human GT is shown on a 1--5 scale.

\subsection{Model Prompts and Configuration}
\label{sec:suppl_prompts}

To ensure reproducibility, we provide the exact text prompts used for training and evaluating FoodSense-VL, as well as for all closed-source and open-source baseline inferences. 

During the two-stage QLoRA fine-tuning, FoodSense-VL is conditioned on the following \textbf{System Prompt} to adopt the persona of a sensory analysis expert:

\begin{tcolorbox}[colback=purple!5, colframe=purple!50, title={\scriptsize\textbf{System Prompt}}, left=4pt, right=4pt, top=4pt, bottom=4pt, boxrule=0.5pt, breakable]
\small\texttt{You are a sensory analysis expert trained to evaluate food images across multiple sensory dimensions. Your task is to analyze the food image and predict sensory experiences based on visual cues.}\\
\\
\texttt{Rate each of these four sensory attributes on a 1.0-5.0 scale: Taste, Smell, Texture, and Sound (when eating). Each sense may have a different rating; do not default to identical ratings across all senses.}\\
\\
\texttt{If reference reviews are provided, treat them as background about similar dishes; do not quote them verbatim and do not mention Yelp.}\\
\\
\texttt{Respond in plain text only (no markdown).}
\end{tcolorbox}

For zero-shot baseline evaluation across generalist VLMs (InternVL, Qwen, LLaVA, Food-Llama, etc.), we utilize the \textbf{User Evaluation Prompt} to extract uniform Sensory Assessment formatting:

\begin{tcolorbox}[colback=gray!5, colframe=gray!50, title={\scriptsize\textbf{User Evaluation Prompt}}, left=4pt, right=4pt, top=4pt, bottom=4pt, boxrule=0.5pt, breakable]
\small\texttt{Analyze this food image and estimate the likely sensory experience. Plain text only (no markdown; no bullets; no asterisks). Provide a Sensory Assessment with 4 sections, one each for Taste, Smell, Texture, and Sound, formatted as: Sense (X.X/5.0): 3-4 sentences of detailed visual justification.}
\end{tcolorbox}

\tcbset{breakable, before skip=4.5pt, after skip=3pt}

\subsection{Sandwich (0001)}
\label{sec:suppl_sandwich}
{\small Human GT: Taste=4.1, Smell=4.1, Texture=4.3, Sound=3.6}

\begin{figure}[H]
\centering
\includegraphics[width=0.6\columnwidth]{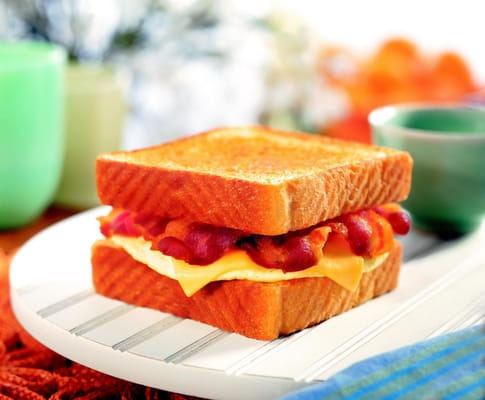}
\end{figure}

\vspace{-6pt}

\scriptsize
\begin{tcolorbox}[halign=flush left, colback=green!5, colframe=green!40, title={\scriptsize\textbf{FoodSense-VL (Stage~2)} --- Taste:3.5\, Smell:3.0\, Texture:2.5\, Sound:2.0}, left=2pt, right=2pt, top=1pt, bottom=1pt, boxrule=0.3pt]
\textbf{Taste:} The sandwich appears \underline{savory} and comforting, with \underline{meaty} notes from the filling and \underline{richness} from \underline{melted} cheese. \underline{Toasted} bread contributes a \underline{mild} \underline{sweet-toasty} balance.\\
\textbf{Smell:} Expected \underline{aroma} is \underline{warm} and inviting, driven by \underline{toasted} bread, cooked egg, processed meat, and \underline{dairy} notes.\\
\textbf{Texture:} The bite likely skews \underline{soft} overall: yielding bread, \underline{melted} cheese, and \underline{smooth} egg with only limited \underline{crispness}.\\
\textbf{Sound:} Predicted sound profile is \underline{quiet}, with \underline{minimal crunch} and mostly \underline{muted chew}.
\end{tcolorbox}

\begin{tcolorbox}[halign=flush left, colback=blue!5, colframe=blue!40, title={\scriptsize\textbf{Base (Gemma~3 27B)} --- Taste:3.5\, Smell:3.0\, Texture:2.5\, Sound:1.5}, left=2pt, right=2pt, top=1pt, bottom=1pt, boxrule=0.3pt]
\textbf{Taste:} The sandwich appears to contain classic breakfast flavors with eggs, cheese, and processed meat, likely sausage. The \underline{toasted} bread suggests slight \underline{caramelized sweetness} that balances \underline{savory} elements. Overall it likely tastes \underline{salty}, \underline{umami-rich}, and \underline{rich} from cheese and egg.\\
\textbf{Smell:} The image suggests a comforting breakfast \underline{aroma}. Expected notes include \underline{cooked egg}, slightly \underline{smoky} processed meat, and \underline{warm}, \underline{wheaty} \underline{toasted bread}, with possible \underline{buttery} hints.\\
\textbf{Texture:} Bread appears \underline{soft} with some structure from toasting. Egg looks \underline{smooth}, while sausage likely adds slight \underline{snap} or \underline{chew}. \underline{Melted} cheese contributes a \underline{stretchy}, yielding mouthfeel with limited contrast.\\
\textbf{Sound:} Eating this would be mostly \underline{quiet}. The \underline{soft} bread and egg produce little \underline{crunch}; likely cues are slight \underline{squish} and a subtle \underline{snap} from processed meat.
\end{tcolorbox}

\begin{tcolorbox}[halign=flush left, colback=gray!5, colframe=gray!50, title={\scriptsize\textbf{InternVL-2.5~26B} --- Taste:4.5\, Smell:4.0\, Texture:4.0\, Sound:3.5}, left=2pt, right=2pt, top=1pt, bottom=1pt, boxrule=0.3pt]
\textbf{Taste:} The sandwich likely offers a \underline{rich}, \underline{savory} taste with the \underline{crispy} bacon providing a \underline{salty}, \underline{smoky} flavor, complemented by the \underline{creamy}, slightly \underline{tangy} taste of \underline{melted} cheese. The \underline{toasted} bread adds a \underline{warm}, slightly \underline{sweet} and \underline{nutty} flavor, enhancing the overall taste experience.\\
\textbf{Smell:} The \underline{aroma} of the sandwich is likely to be enticing, with the scent of \underline{crispy} bacon and \underline{melted} cheese dominating. The \underline{toasted} bread adds a \underline{subtle}, comforting \underline{aroma} that complements the other flavors, making it a delightful sensory experience.\\
\textbf{Texture:} The sandwich offers a delightful contrast in textures, with the \underline{crispy} bacon providing a satisfying \underline{crunch}, the \underline{melted} cheese offering a \underline{smooth}, \underline{gooey} texture, and the \underline{toasted} bread adding a \underline{firm}, slightly \underline{chewy} base. The combination creates a \underline{mouthfeel} that is both satisfying and enjoyable.\\
\textbf{Sound:} While the visual appeal is strong, the sound of this sandwich is less pronounced. The \underline{crunch} of the bacon and the slight \underline{sizzle} of the cheese might be audible, but the overall sound is likely to be minimal, focusing more on the visual and olfactory senses.
\end{tcolorbox}

\begin{tcolorbox}[halign=flush left, colback=brown!5, colframe=brown!50, title={\scriptsize\textbf{Food-Llama~11B} --- Taste:--\, Smell:--\, Texture:--\, Sound:--}, left=2pt, right=2pt, top=1pt, bottom=1pt, boxrule=0.3pt]
\textbf{Taste:} The generated output describes a hearty \underline{savory} profile from \underline{melted} cheese, \underline{toasted} bread, and bacon-like filling.\\
\textbf{Smell:} It suggests a \underline{warm} \underline{cooked aroma} associated with bacon, \underline{toasted} bread, and \underline{rich dairy} elements.\\
\textbf{Texture:} It emphasizes a comfort-food texture blend with \underline{crisp} toast edges and \underline{creamy} \underline{melted} components.\\
\textbf{Sound:} The output references \underline{mild sizzling}/\underline{crispness} cues rather than loud \underline{crunch}.
\end{tcolorbox}

\begin{tcolorbox}[halign=flush left, colback=purple!5, colframe=purple!50, title={\scriptsize\textbf{Qwen2.5-VL~32B} --- Taste:4.5\, Smell:4.0\, Texture:4.5\, Sound:2.5}, left=2pt, right=2pt, top=1pt, bottom=1pt, boxrule=0.3pt]
\textbf{Taste:} The sandwich appears to be filled with \underline{melted} cheese, which suggests a \underline{creamy} and slightly \underline{salty} flavor. The presence of what looks like bacon or a similar ingredient indicates a \underline{savory}, \underline{smoky} taste that would complement the cheese well. The \underline{toasted} bread adds a \underline{subtle sweetness} and a hint of \underline{crunch}, enhancing the overall \underline{richness} of the sandwich. The combination of these elements likely results in a \underline{balanced}, satisfying, and \underline{indulgent} taste profile.\\
\textbf{Smell:} The image shows a freshly \underline{toasted} sandwich, which implies a \underline{warm}, inviting \underline{aroma} of golden-brown bread. The \underline{melted} cheese would contribute a \underline{rich}, \underline{buttery} scent, while the bacon or similar filling might add a \underline{savory}, slightly \underline{smoky} odor. The overall smell is likely to be comforting and appetizing, evoking a sense of warmth and homeliness.\\
\textbf{Texture:} The \underline{toasted} bread has a \underline{crispy} exterior and a \underline{soft} interior, providing a pleasing contrast in texture. The \underline{melted} cheese appears \underline{smooth} and \underline{gooey}, offering a \underline{creamy} texture that would \underline{melt} in the mouth. The bacon or similar filling seems to have a slightly \underline{chewy}, \underline{crispy} texture, adding another layer of variety. The combination of these textures creates a satisfying and diverse \underline{mouthfeel}.\\
\textbf{Sound:} While the image itself does not provide auditory information, one can imagine the sound of biting into the sandwich. The \underline{crispy} toast would likely produce a satisfying \underline{crunch}, while the cheese might stretch slightly, creating a gentle tearing sound. The overall sound experience would be minimal but could include the \underline{soft squish} of the cheese and the slight resistance from the filling, contributing to an engaging eating experience. However, without actual audio, this assessment is speculative.
\end{tcolorbox}

\begin{tcolorbox}[halign=flush left, colback=black!5, colframe=black!40, title={\scriptsize\textbf{GPT~5.4~Thinking} --- Taste:4.2\, Smell:3.9\, Texture:4.3\, Sound:3.4}, left=2pt, right=2pt, top=1pt, bottom=1pt, boxrule=0.3pt]
\textbf{Taste:} This looks strongly \underline{savory}, \underline{rich}, and slightly \underline{salty}, with the \underline{toasted} bread, \underline{melted} cheese, and reddish meat filling suggesting a classic comfort-food profile. The glossy red sauce adds a likely \underline{sweet-tangy} note that would brighten the heavier cheese and meat. Visually, the sandwich appears \underline{indulgent} rather than \underline{subtle}, with bold, familiar flavors layered closely together. Overall, it seems likely to taste \underline{warm}, \underline{hearty}, and satisfying, with mild \underline{sweetness} balancing the fat and salt.\\
\textbf{Smell:} The image suggests a \underline{warm aroma} led by \underline{toasted bread} and \underline{melted cheese}, which usually create a \underline{nutty}, \underline{buttery} scent when heated. The meat filling likely contributes a \underline{smoky}, cured, or \underline{grilled} smell that would make the sandwich feel more substantial. The red sauce probably adds a faint \underline{sweet-acidic} note, similar to ketchup or barbecue sauce, giving the aroma a \underline{sharper} edge. Altogether, it seems like a noticeable but not highly \underline{complex} smell, more \underline{cozy} and appetizing than intensely fragrant.\\
\textbf{Texture:} The bread looks well \underline{toasted} on the outside, so the first bite would probably have a light \underline{crispness} before giving way to a \underline{soft} interior crumb. The cheese appears \underline{warm} and pliable, adding a \underline{smooth}, slightly \underline{stretchy} layer between the \underline{firmer} bread slices. The filling looks \underline{juicy} and a bit \underline{bulky}, which suggests a \underline{dense}, \underline{meaty} bite with some \underline{softness} from the sauce. Overall, the texture likely combines \underline{crunch}, \underline{softness}, and \underline{gooeyness} in a very tactile, layered way.\\
\textbf{Sound:} This sandwich looks like it would make a gentle \underline{crisp} bite sound rather than a loud \underline{crackle}, because the bread is toasted but still fairly thick and \underline{soft} inside. The edges may give a light \underline{crunch} when bitten, especially at the corners where the toast looks \underline{drier} and \underline{firmer}. The cheese and \underline{saucy} filling would likely \underline{mute} the sound quickly, turning the bite into more of a \underline{soft chew} than a \underline{brittle snap}. So the eating sound would probably be a brief crunch followed by a quiet, \underline{dense}, slightly \underline{squishy} chew.
\end{tcolorbox}

\begin{tcolorbox}[halign=flush left, colback=cyan!5, colframe=cyan!40, title={\scriptsize\textbf{Gemini~3.1~Pro} --- Taste:4.2\, Smell:3.8\, Texture:4.5\, Sound:1.2}, left=2pt, right=2pt, top=1pt, bottom=1pt, boxrule=0.3pt]
\textbf{Taste:} The vibrant green powder covering the entire top surface strongly indicates a rich, \underline{earthy}, and slightly \underline{bitter} matcha flavor. Beneath this top layer, the pale \underline{creamy} section suggests a contrasting \underline{sweet} and \underline{milky} profile, likely balancing the bitterness of the tea. The overall taste profile would be a harmonious blend of \underline{sweet dairy} and \underline{astringent}, \underline{herbaceous} green tea.\\
\textbf{Smell:} The thick dusting of matcha powder will provide a \underline{dominant}, \underline{earthy}, and \underline{grassy aroma} that is characteristic of high-quality green tea. The creamy layer underneath might offer subtle, \underline{sweet dairy} or \underline{vanilla} notes that soften the intense tea scent. Together, these visual cues suggest a \underline{fragrant}, \underline{botanical}, and \underline{comforting} smell that is not overly sweet.\\
\textbf{Texture:} The fine, \underline{dry} powder on top will provide a slightly \underline{velvety} and \underline{dusty} initial mouthfeel. This will quickly give way to the \underline{smooth}, \underline{rich}, and \underline{aerated} texture of the cream or mousse layer visible just below the rim. The contrast between the \underline{dry}, \underline{powdery} topping and the \underline{lush}, \underline{creamy} base suggests a \underline{delicate} and \underline{melting} consistency on the palate.\\
\textbf{Sound:} The \underline{soft}, yielding nature of the powdered topping and the creamy mousse indicates that eating this dessert would produce almost no audible sound. The wooden spoon resting on the saucer might make a \underline{dull}, \underline{gentle clink} when picking up a portion, but the food itself is entirely soft. Therefore, any sound would be limited to the \underline{quiet}, \underline{muted} noise of the spoon sinking into the delicate layers.
\end{tcolorbox}

\begin{tcolorbox}[halign=flush left, colback=red!5, colframe=red!50, title={\scriptsize\textbf{LLaVA} --- Taste:3.5\, Smell:4.0\, Texture:3.0\, Sound:3.0}, left=2pt, right=2pt, top=1pt, bottom=1pt, boxrule=0.3pt]
\textbf{Taste:} The sandwich is described as \underline{savory} and \underline{rich}, with \underline{toasted} bread, \underline{melted} cheese, and \underline{meaty} filling as primary cues.\\
\textbf{Smell:} \underline{Aroma} is inferred as \underline{warm} and appetizing from toasted bread, dairy, and cooked protein.\\
\textbf{Texture:} Predicted texture combines \underline{soft} interior components with modest toast \underline{crispness}.\\
\textbf{Sound:} Expected sound remains \underline{moderate-to-quiet}, with occasional \underline{soft crunch}.
\end{tcolorbox}

\begin{tcolorbox}[halign=flush left, colback=orange!5, colframe=orange!50, title={\scriptsize\textbf{Claude~4.6~Opus} --- Taste:4.2\, Smell:3.9\, Texture:4.0\, Sound:3.5}, left=2pt, right=2pt, top=1pt, bottom=1pt, boxrule=0.3pt]
\textbf{Taste:} This appears to be a grilled cheese and bacon sandwich made with golden-toasted white bread and what looks like melted American or cheddar cheese layered with strips of crispy bacon. The taste profile would be dominated by rich, \underline{savory}, \underline{salty}, and \underline{umami} notes from the rendered bacon fat and melted cheese, complemented by a mild \underline{buttery sweetness} from the toasted bread. The combination of salty cured pork and creamy melted cheese creates a deeply satisfying, \underline{indulgent} flavor that registers high on the palatability scale. A slight caramelization on the bread surface suggests a Maillard reaction that would add \underline{toasty}, \underline{nutty} depth to the overall flavor.\\
\textbf{Smell:} The aroma emanating from this sandwich would be warm and immediately appetizing, led by the \underline{smoky}, fatty scent of cooked bacon which tends to carry strongly through the air. Melted cheese contributes a rich, slightly \underline{tangy} dairy note that blends with the toasty, yeasty fragrance of the grilled bread. The overall olfactory impression would be one of comfort food, with the rendered bacon grease and butter used for toasting creating a \underline{savory}, slightly smoky haze. The scent would not be particularly complex or layered compared to dishes with herbs or spices, but it would be intensely appetizing and warm.\\
\textbf{Texture:} The exterior of the bread shows a deep golden-brown crust that indicates a firm, \underline{crunchy} bite on the outside, while the interior crumb would remain \underline{soft} and slightly compressed from the grilling process. The melted cheese visible between the slices appears \underline{gooey} and \underline{stretchy}, providing a smooth, creamy contrast to the crisp bread. The bacon strips look like they were cooked to a moderately \underline{crispy} state, adding a \underline{snappy}, slightly \underline{chewy} element that breaks up the softness of the cheese and bread. The interplay between the three distinct textures, crunchy toast, molten cheese, and crisp-chewy bacon, creates a satisfying and varied mouthfeel.\\
\textbf{Sound:} Biting into this sandwich would produce an audible \underline{crunch} from the well-toasted bread exterior, which shows enough browning and firmness to generate a satisfying initial snap. The bacon would contribute additional \underline{crackling} sounds as teeth break through the rendered strips, adding a secondary layer of crispness. Once past the outer crust, the sound would diminish as the softer cheese and inner bread muffle the bite, making the acoustic experience front-loaded. Cutting the sandwich in half might produce a slight scraping sound against the plate, and pulling the halves apart would create a quiet stretching sound from the melted cheese strands.
\end{tcolorbox}


\clearpage

\subsection{Steak Rice (0002)}
\label{sec:suppl_steak}
{\small Human GT: Taste=4.3, Smell=4.3, Texture=4.4, Sound=4.1}

\begin{figure}[H]
\centering
\includegraphics[width=0.6\columnwidth]{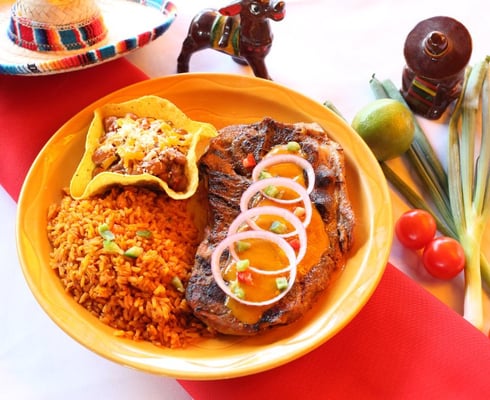}
\end{figure}

\vspace{-6pt}

\scriptsize
\begin{tcolorbox}[halign=flush left, colback=green!5, colframe=green!40, title={\scriptsize\textbf{FoodSense-VL (Stage~2)} --- Taste:4.0\, Smell:3.5\, Texture:3.0\, Sound:2.5}, left=2pt, right=2pt, top=1pt, bottom=1pt, boxrule=0.3pt]
\textbf{Taste:} The deep, dark color of the steak suggests a \underline{robust}, \underline{savory} flavor with a slightly \underline{charred} tone. The red rice and visible garnish imply a \underline{spicy}, tomato-forward base. Onion rings add \underline{pungency} and contrast, and the plated components suggest a \underline{bold}, \underline{well-seasoned} profile.\\
\textbf{Smell:} Grill marks and sheen on the steak suggest a \underline{sizzling}, \underline{savory aroma}. The rice color cues \underline{warm spice} notes, while lime and green onion suggest a \underline{bright citrus} lift. Overall smell appears \underline{rich} and layered.\\
\textbf{Texture:} The steak looks \underline{firm} and \underline{chewy}, with \underline{fluffy} separated rice grains. The tortilla/chip element appears \underline{brittle} and \underline{crunchy}, while onion rings add slight bite resistance.\\
\textbf{Sound:} The dish likely stays relatively \underline{quiet} overall, with most sound from a \underline{crisp snap} in the tortilla/chip element and subtle \underline{rustling} from rice.
\end{tcolorbox}

\begin{tcolorbox}[halign=flush left, colback=purple!5, colframe=purple!50, title={\scriptsize\textbf{Qwen2.5-VL~32B} --- Taste:4.5\, Smell:4.0\, Texture:4.0\, Sound:2.5}, left=2pt, right=2pt, top=1pt, bottom=1pt, boxrule=0.3pt]
\textbf{Taste:} The dish appears \underline{rich} and \underline{flavorful}, with \underline{savory}, slightly \underline{smoky} grilled meat, \underline{warm earthy} orange rice, and a \underline{creamy} side component. Pickled onions add a \underline{tangy}, \underline{sharp} contrast.\\
\textbf{Smell:} The profile is likely \underline{robust}, led by the \underline{charred aroma} of grilled meat and \underline{fragrant spices}. Lime and fresh vegetables suggest a \underline{citrusy} lift, while pickled onions add a \underline{pungent vinegary} note.\\
\textbf{Texture:} Expected texture is \underline{varied}: \underline{tender-juicy} meat with a slightly \underline{crispy charred} exterior, \underline{fluffy moist} rice, a \underline{creamy-smooth} side, and \underline{crisp crunchy} onion accents.\\
\textbf{Sound:} Auditory cues are mostly \underline{subtle}, with imagined \underline{sizzle} from grilled meat, occasional \underline{crunch} from onions, and \underline{soft chew} from rice and side components.
\end{tcolorbox}

\begin{tcolorbox}[halign=flush left, colback=blue!5, colframe=blue!40, title={\scriptsize\textbf{Base (Gemma~3 27B)} --- Taste:3.8\, Smell:3.5\, Texture:3.2\, Sound:2.5}, left=2pt, right=2pt, top=1pt, bottom=1pt, boxrule=0.3pt]
\textbf{Taste:} The steak likely has a \underline{savory}, \underline{umami-rich} profile with hints of \underline{char} from grilling. The red rice suggests tomato-based seasoning with slight \underline{acidity} and \underline{sweetness}. Lime and pickled onions add a \underline{fresh}, \underline{tangy} counterpoint that balances the richer components.\\
\textbf{Smell:} Expected \underline{aroma} combines \underline{smoky}, \underline{meaty} grilled steak with \underline{warm} spice notes from rice. Lime adds \underline{bright citrus}, while onions add a \underline{pungent}, \underline{sharp} top note.\\
\textbf{Texture:} Steak appears \underline{seared} outside and \underline{tender}, \underline{juicy} inside. Rice looks \underline{fluffy}, chips provide distinct \underline{crunch}, and onions contribute \underline{crisp} bite with occasional \underline{juicy} bursts from lime.\\
\textbf{Sound:} The dish should be \underline{moderately audible}: chips deliver \underline{loud crunch}, steak adds \underline{soft chew}, while rice and garnish remain mostly \underline{quiet}.
\end{tcolorbox}

\begin{tcolorbox}[halign=flush left, colback=gray!5, colframe=gray!50, title={\scriptsize\textbf{InternVL-2.5~26B} --- Taste:4.5\, Smell:4.0\, Texture:4.0\, Sound:3.5}, left=2pt, right=2pt, top=1pt, bottom=1pt, boxrule=0.3pt]
\textbf{Taste:} The dish likely offers a \underline{rich} and diverse flavor profile. The \underline{grilled} steak would provide a \underline{savory}, slightly \underline{charred} taste, complemented by the \underline{tangy}, \underline{creamy} sauce. The rice, \underline{seasoned} with spices, would add a \underline{warm}, slightly \underline{spicy} note, while the taco would offer a mix of \underline{savory} meat and a hint of cheese. The lime would add a \underline{fresh}, \underline{citrusy} zing to balance the flavors.\\
\textbf{Smell:} The \underline{aroma} would be inviting, with the \underline{smoky} scent of the \underline{grilled} steak mingling with the \underline{earthy}, slightly \underline{spicy} \underline{aroma} of the rice. The \underline{fresh} lime and herbs would add a \underline{bright}, \underline{citrusy} note, enhancing the overall olfactory experience.\\
\textbf{Texture:} The dish would offer a variety of textures. The steak would be \underline{tender} and \underline{juicy}, contrasting with the \underline{crispy}, slightly \underline{crunchy} taco shell. The rice would be \underline{fluffy} and slightly \underline{sticky}, while the sauce would add a \underline{smooth}, \underline{creamy} element. The raw vegetables would provide a \underline{crisp}, refreshing contrast.\\
\textbf{Sound:} The sound of this dish would primarily be the \underline{sizzle} of the steak as it cooks, followed by the gentle \underline{clinking} of utensils against the plate. The \underline{crunch} of the taco shell and the \underline{soft rustle} of the lime and herbs would add subtle auditory details, enhancing the overall sensory experience.
\end{tcolorbox}

\begin{tcolorbox}[halign=flush left, colback=black!5, colframe=black!40, title={\scriptsize\textbf{GPT~5.4~Thinking} --- Taste:4.6\, Smell:4.4\, Texture:4.4\, Sound:3.7}, left=2pt, right=2pt, top=1pt, bottom=1pt, boxrule=0.3pt]
\textbf{Taste:} This dish looks intensely \underline{savory} and \underline{spiced}, with \underline{grilled meat}, \underline{seasoned rice}, and a filled tortilla shell suggesting a \underline{bold}, \underline{layered} flavor profile. The char on the meat implies \underline{smoky}, slightly \underline{bitter} edges that would deepen the \underline{richness}, while the onions and salsa-like garnish likely add \underline{brightness} and a mild \underline{acidic lift}. The \underline{reddish rice} appears well seasoned, hinting at tomato, garlic, and spice, which would make the plate feel \underline{hearty} and \underline{robust}. Overall, it seems likely to taste \underline{rich}, \underline{smoky}, \underline{salty}, and \underline{warmly spiced}, with small bursts of \underline{freshness} from the toppings.\\
\textbf{Smell:} The strongest visual cue is the grilled meat, which suggests a pronounced \underline{smoky}, \underline{roasted aroma} coming off the dark \underline{seared} surface. The seasoned rice and stuffed shell likely contribute \underline{warm} smells of spices, oil, and cooked starch, creating a \underline{full} and \underline{comforting} scent. Fresh onion, lime, and nearby vegetables imply a lighter \underline{sharpness} that could cut through the heavier grilled notes. Altogether, the aroma seems \underline{vivid} and appetizing, combining smoke, spice, and a \underline{fresh acidic edge}.\\
\textbf{Texture:} The meat looks \underline{firm} and \underline{juicy}, with a \underline{charred exterior} that would give some resistance before yielding to a \underline{tender} interior. The rice appears \underline{fluffy} yet \underline{oily} enough to cling together slightly, creating a \underline{soft} and \underline{granular mouthfeel}. The stuffed tortilla shell seems \underline{crisp} at the edges but \underline{softened} by its filling, adding a contrast between \underline{crunch} and \underline{softness}. Overall, the plate likely offers a satisfying mix of \underline{chew}, \underline{fluffiness}, \underline{crisp edges}, and \underline{moist}, \underline{juicy} bites.\\
\textbf{Sound:} This meal seems \underline{moderately audible}, mostly through the slight \underline{scrape} and bite resistance of the grilled meat rather than any loud \underline{crunch}. The \underline{charred exterior} of the meat and the \underline{crisp shell edges} could produce a faint \underline{crackle} or \underline{crisp snap} when cut or bitten. The rice and toppings would be much \underline{quieter}, contributing \underline{soft}, \underline{muted} eating sounds. So the overall sound profile would likely be a combination of \underline{light crispness} and steady, substantial \underline{chewing} rather than sharp crunches.
\end{tcolorbox}

\begin{tcolorbox}[halign=flush left, colback=cyan!5, colframe=cyan!40, title={\scriptsize\textbf{Gemini~3.1~Pro} --- Taste:4.5\, Smell:4.6\, Texture:4.3\, Sound:2.8}, left=2pt, right=2pt, top=1pt, bottom=1pt, boxrule=0.3pt]
\textbf{Taste:} The dark, \underline{glossy glaze} coating the skewers suggests a \underline{rich}, deeply \underline{savory}, and slightly \underline{sweet} flavor profile characteristic of teriyaki or tare sauce. Prominent \underline{charred edges} on the meat pieces indicate a \underline{robust}, \underline{bitter}, and \underline{smoky} undertone that balances the sweetness. The presence of a \underline{fresh} lemon wedge implies an optional burst of sharp \underline{acidity} to cut through the heavy \underline{umami} flavors.\\
\textbf{Smell:} The heavily \underline{blistered} and \underline{blackened} edges of the meat point to a \underline{strong}, \underline{intense smoky aroma} from high-heat grilling. The thick, \underline{caramelized} coating will contribute a \underline{fragrant}, \underline{savory} scent of roasted soy and caramelized sugars. An underlying hint of \underline{bright}, \underline{fresh citrus} will mingle in the air if the adjacent lemon is utilized.\\
\textbf{Texture:} The \underline{sticky}, \underline{glistening} sauce promises a \underline{tacky} and \underline{clinging mouthfeel} upon the first bite. This is immediately contrasted by the \underline{crispy}, \underline{rigid} texture of the charred exterior crust gripping the meat. The distinct, thick chunks threaded onto the wooden sticks appear \underline{plump} and \underline{moist}, suggesting a \underline{tender}, \underline{yielding} interior beneath the slight exterior \underline{crunch}.\\
\textbf{Sound:} Removing the \underline{sticky} chunks of meat from the wooden skewers will likely produce a subtle, \underline{damp sliding noise}. The first bite will yield a faint but distinct \underline{crunching sound} due to the crystallized glaze and charred edges. Subsequent chewing will be relatively \underline{quiet} and \underline{soft}, dominated by the \underline{tender} and \underline{juicy} interior of the grilled meat.
\end{tcolorbox}

\begin{tcolorbox}[halign=flush left, colback=red!5, colframe=red!50, title={\scriptsize\textbf{LLaVA} --- Taste:3.5\, Smell:4.0\, Texture:3.0\, Sound:3.0}, left=2pt, right=2pt, top=1pt, bottom=1pt, boxrule=0.3pt]
\textbf{Taste:} The dish appears Mexican-inspired, with \underline{savory} meat, sauce, and rice as the core flavor base. Onion and tomato elements suggest \underline{fresh tang} and contrast.\\
\textbf{Smell:} Likely \underline{aromatic} and \underline{savory}, with grilled-meat notes and \underline{spiced} sauce. Rice adds a \underline{mild earthy} background smell.\\
\textbf{Texture:} Expected texture mix is \underline{tender} meat, \underline{soft} rice, and some \underline{chew} from tortilla components, with occasional \underline{crispness} from garnish.\\
\textbf{Sound:} Predominantly \underline{gentle} eating sounds, with occasional \underline{crunch} from tortilla elements and \underline{soft} utensil contact.
\end{tcolorbox}

\begin{tcolorbox}[halign=flush left, colback=orange!5, colframe=orange!50, title={\scriptsize\textbf{Claude~4.6~Opus} --- Taste:4.5\, Smell:4.3\, Texture:4.2\, Sound:3.2}, left=2pt, right=2pt, top=1pt, bottom=1pt, boxrule=0.3pt]
\textbf{Taste:} This is a Mexican-style plate featuring a large \underline{grilled} or \underline{pan-seared} steak topped with sliced red onion rings and what appears to be a \underline{warm} salsa or ranchero sauce, alongside a portion of Mexican rice and a taco or tostada with crumbled cheese. The steak shows deep char marks and dark browning across its surface, indicating intense Maillard caramelization that would deliver \underline{bold}, \underline{smoky}, \underline{beefy umami} flavors. The rice has a vivid orange-red hue suggesting it was cooked with tomato, cumin, and possibly chili, contributing a \underline{savory}, mildly \underline{spicy}, and slightly \underline{tangy} base note. The layered composition of \underline{rich} grilled meat, \underline{seasoned} rice, and a cheese-topped taco creates a deeply satisfying flavor profile with interplay between \underline{smoky}, \underline{salty}, \underline{tangy}, and mildly \underline{spicy} elements.\\
\textbf{Smell:} The dominant \underline{aroma} would be the \underline{smoky}, \underline{charred} scent of the grilled steak, which carries powerfully and is one of the most universally appetizing smells in cooking. Underneath that, the Mexican rice would contribute a \underline{warm}, \underline{toasty} fragrance with notes of \underline{cumin}, \underline{garlic}, and \underline{tomato} that are characteristic of well-seasoned arroz rojo. The raw red onion rings and the salsa draped over the meat would add a \underline{sharp}, slightly \underline{pungent}, and \underline{bright acidic} note that cuts through the heavier meat and fat aromas. Fresh ingredients visible nearby on the table, including cherry tomatoes, green onions, and a lime, suggest garnishing elements whose bright, \underline{herbaceous}, and \underline{citrus} scents would round out the olfactory experience with freshness.\\
\textbf{Texture:} The steak appears to have a \underline{firm}, well-seared crust on the exterior from the grill marks, which would give way to a denser, \underline{chewier} interior consistent with a thicker cut cooked to at least medium doneness based on the degree of surface char. The Mexican rice looks \underline{fluffy} yet slightly \underline{sticky}, with individual grains holding together in \underline{soft} clumps that would provide a \underline{pillowy} contrast to the meat. The taco or tostada shell has a visible corn tortilla base that would contribute a light \underline{crunch} or \underline{soft} pliability depending on whether it was fried or simply warmed, and the crumbled cheese on top adds a \underline{crumbly}, slightly \underline{grainy} texture. The sliced onion rings on the steak remain raw and \underline{crisp}, providing a \underline{snappy}, \underline{juicy} bite that contrasts sharply with the cooked elements.\\
\textbf{Sound:} Cutting into the steak with a knife would produce a \underline{firm resistance} and a light \underline{scraping sound} against the plate, and the seared crust might yield a faint \underline{crackling} as the blade breaks through the charred exterior. Chewing the rice would be relatively \underline{quiet}, producing \underline{soft}, \underline{muffled} sounds typical of cooked grains, contributing \underline{minimal acoustic} presence to the eating experience. The raw onion rings would offer a mild \underline{snap} and a juicy \underline{crunch} when bitten, adding a brief audible moment to each bite that includes them. If the taco shell was lightly fried or toasted, biting into it would generate a \underline{moderate crunch}, though this would be somewhat \underline{dampened} by the \underline{moist} fillings and melted cheese on top.
\end{tcolorbox}

\begin{tcolorbox}[halign=flush left, colback=brown!5, colframe=brown!50, title={\scriptsize\textbf{Food-Llama~11B} --- Taste:4.0\, Smell:4.0\, Texture:4.0\, Sound:3.0}, left=2pt, right=2pt, top=1pt, bottom=1pt, boxrule=0.3pt]
\textbf{Taste:} The plate suggests a \underline{rich} \underline{savory} profile from \underline{grilled} steak and \underline{seasoned} rice, with \underline{tang} from onion/lime garnish and \underline{creamy} contrast from cheese-covered tortilla.\\
\textbf{Smell:} Likely \underline{smoky} and \underline{warm} from the \underline{grilled} meat, with a \underline{spiced} rice \underline{aroma} and \underline{fresh citrus-herb} top notes.\\
\textbf{Texture:} Expected texture is mixed: \underline{tender-chewy} steak, \underline{fluffy} rice, \underline{crunchy} tortilla/chip components, and \underline{creamy-soft} topping elements.\\
\textbf{Sound:} Audible cues are \underline{moderate}, mainly from \underline{crisp tortilla crunch} and light utensil sounds, with \underline{softer chew} from meat and rice.
\end{tcolorbox}

\subsection{Southern Scampi (0010)}
\label{sec:suppl_scampi}
{\small Human GT: Taste=3.7, Smell=3.5, Texture=3.5, Sound=2.7}

\begin{figure}[H]
\centering
\includegraphics[width=0.6\columnwidth]{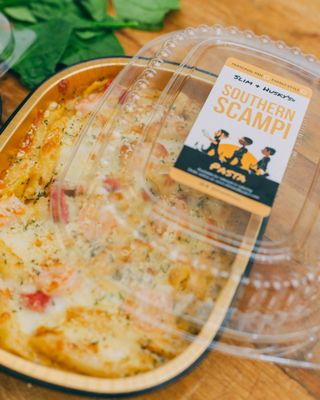}
\end{figure}

\vspace{-6pt}

\scriptsize
\begin{tcolorbox}[halign=flush left, colback=green!5, colframe=green!40, title={\scriptsize\textbf{FoodSense-VL (Stage~2)} --- Taste:3.5\, Smell:3.0\, Texture:2.5\, Sound:2.0}, left=2pt, right=2pt, top=1pt, bottom=1pt, boxrule=0.3pt]
\textbf{Taste:} The baked scampi pasta appears \underline{savory} and \underline{creamy}, with \underline{seafood} notes from shrimp and \underline{mild acidity} from tomato pieces.\\
\textbf{Smell:} Predicted \underline{aroma} is \underline{warm} and \underline{dairy-forward}, with \underline{cooked seafood} and \underline{herb} accents.\\
\textbf{Texture:} The dish looks mostly \underline{soft} and yielding: \underline{creamy} sauce, \underline{tender} pasta, and small shrimp pieces with limited \underline{crunch}.\\
\textbf{Sound:} Overall sound is expected to be \underline{quiet}, with \underline{soft} utensil and \underline{chew} cues rather than sharp \underline{crackle}.
\end{tcolorbox}

\begin{tcolorbox}[halign=flush left, colback=blue!5, colframe=blue!40, title={\scriptsize\textbf{Base (Gemma~3 27B)} --- Taste:3.8\, Smell:3.5\, Texture:3.2\, Sound:2.0}, left=2pt, right=2pt, top=1pt, bottom=1pt, boxrule=0.3pt]
\textbf{Taste:} Visible shrimp, pasta, and \underline{creamy}, \underline{cheesy} sauce suggest a \underline{rich}, \underline{savory} profile. Herbs and tomato pieces likely add mild \underline{acidity} that cuts richness. The scampi label implies \underline{buttery}, \underline{garlic-forward} flavor with possible mild \underline{spice}.\\
\textbf{Smell:} The dish likely smells of cooked seafood and \underline{dairy} with \underline{herb} notes. \underline{Browned cheese} suggests a slightly \underline{caramelized}, \underline{nutty} scent. A \underline{buttery} and subtle \underline{garlic} \underline{aroma} is expected.\\
\textbf{Texture:} Likely \underline{layered}: \underline{soft} pasta, shrimp with slight \underline{bite}, and \underline{creamy}, \underline{smooth} sauce. The top may have mild \underline{crisp} edges, but overall the dish appears \underline{soft} and yielding.\\
\textbf{Sound:} Expected minimal sound during eating, mostly utensil contact with container and \underline{soft chew}; little \underline{crunch} or \underline{snap}.
\end{tcolorbox}

\begin{tcolorbox}[halign=flush left, colback=gray!5, colframe=gray!50, title={\scriptsize\textbf{InternVL-2.5~26B} --- Taste:4.5\, Smell:4.0\, Texture:4.0\, Sound:3.5}, left=2pt, right=2pt, top=1pt, bottom=1pt, boxrule=0.3pt]
\textbf{Taste:} The Southern Scampi pasta likely offers a \underline{rich}, \underline{savory} flavor profile with a hint of \underline{garlic} and lemon, complemented by the \underline{brininess} of the shrimp. The \underline{creamy} sauce would add a \underline{smooth}, \underline{indulgent} quality, while the pasta provides a satisfying \underline{chew}.\\
\textbf{Smell:} The \underline{aroma} of the dish would be enticing, with the scent of sauteed \underline{garlic} and \underline{herbs} mingling with the \underline{buttery}, \underline{cheesy} sauce. The shrimp would add a \underline{subtle seafood aroma}, making the dish smell hearty and comforting.\\
\textbf{Texture:} The texture of the pasta would be \underline{al dente}, providing a pleasant resistance to the bite. The \underline{creamy} sauce would coat the pasta evenly, while the shrimp would offer a \underline{tender}, slightly \underline{firm} bite. The cheese on top would be \underline{melted} and \underline{gooey}, adding a contrasting \underline{softness}.\\
\textbf{Sound:} While the dish itself wouldn't make much noise, the act of eating it would likely involve the satisfying sounds of pasta being twirled and the occasional \underline{crunch} of shrimp. The \underline{creamy} sauce might create a slight \underline{slurping sound} as it coats the pasta.
\end{tcolorbox}

\begin{tcolorbox}[halign=flush left, colback=purple!5, colframe=purple!50, title={\scriptsize\textbf{Qwen2.5-VL~32B} --- Taste:4.0\, Smell:4.0\, Texture:4.5\, Sound:2.5}, left=2pt, right=2pt, top=1pt, bottom=1pt, boxrule=0.3pt]
\textbf{Taste:} The image shows a dish labeled "Southern Scampi Pasta," which suggests a \underline{rich} and flavorful meal. The visible \underline{creamy} sauce indicates a likely \underline{buttery} and \underline{savory} taste profile, enhanced by the presence of what appears to be shrimp or scampi, adding a \underline{briny} and slightly \underline{sweet} element. The golden-brown crust on top hints at a \underline{crispy}, \underline{toasted} texture that could contribute a \underline{nutty} flavor when combined with the pasta. The overall appearance suggests a \underline{well-balanced} dish with a mix of \underline{umami}, \underline{salty}, and slightly \underline{sweet} notes.\\
\textbf{Smell:} While the image doesn't convey smell directly, the visual cues suggest an \underline{aromatic} experience. The \underline{creamy} sauce and the golden-brown crust imply a scent of \underline{melted butter}, \underline{garlic}, and possibly \underline{herbs} like parsley or basil, which are common in scampi dishes. The shrimp or scampi would add a \underline{fresh seafood aroma}, while the \underline{toasted} elements might introduce a \underline{warm}, \underline{nutty} fragrance. The combination of these elements suggests a comforting and inviting smell.\\
\textbf{Texture:} The dish presents a variety of textures. The pasta appears \underline{soft} and \underline{al dente}, coated in a \underline{creamy} sauce that looks \underline{smooth} and \underline{velvety}. The golden-brown crust on top suggests a \underline{crunchy}, slightly \underline{crispy} layer, providing contrast to the softer elements. The shrimp or scampi pieces likely have a \underline{tender}, slightly \underline{chewy} texture, complementing the smoothness of the sauce and the firmness of the pasta. The overall texture seems \underline{well-balanced}, offering a satisfying \underline{mouthfeel}.\\
\textbf{Sound:} The image itself does not provide auditory information, but based on the visual cues, one can infer potential sounds associated with eating this dish. The \underline{creamy} sauce might make a slight \underline{slurping} sound as it coats the pasta, while the \underline{crispy} crust could produce a satisfying \underline{crunch} when bitten into. The shrimp or scampi might offer a gentle \underline{sizzle} if freshly cooked, though this is speculative without actual audio. Overall, the sound experience would likely be \underline{subtle} but enjoyable, focusing on the satisfaction of textures rather than loud noises.
\end{tcolorbox}

\begin{tcolorbox}[halign=flush left, colback=black!5, colframe=black!40, title={\scriptsize\textbf{GPT~5.4~Thinking} --- Taste:4.4\, Smell:4.5\, Texture:4.2\, Sound:2.6}, left=2pt, right=2pt, top=1pt, bottom=1pt, boxrule=0.3pt]
\textbf{Taste:} This dish looks \underline{rich}, \underline{savory}, and \underline{buttery}, with the \underline{browned cheese} and \underline{creamy} surface suggesting a deeply \underline{comforting} flavor. The label and appearance imply a shrimp-based preparation, so it likely has a mild \underline{seafood sweetness} underneath garlic, salt, and \underline{herb-heavy seasoning}. The \underline{golden spots} on top suggest \underline{oven-browned edges} that would add a slightly \underline{roasted}, almost \underline{nutty} note. Overall, it seems likely to taste \underline{warm}, \underline{creamy}, and \underline{indulgent}, with a strong \underline{savory depth} and a gentle \underline{briny lift}.\\
\textbf{Smell:} The visual cues point to a pronounced \underline{aroma} of \underline{butter}, \underline{garlic}, and \underline{baked cheese} rising from the hot surface. If the dish is indeed shrimp-based, there would probably also be a light \underline{oceanic} smell, but softened by the \underline{creamy sauce} and herbs rather than \underline{sharp} or \underline{fishy}. The browned top suggests a \underline{toasty} baked scent that makes the whole dish feel \underline{heavier} and more \underline{fragrant}. Altogether, it likely smells \underline{robust}, \underline{cozy}, and highly appetizing, with garlic and butter leading the profile.\\
\textbf{Texture:} The top looks \underline{soft} and \underline{molten}, with a \underline{creamy}, slightly \underline{thick} consistency under a lightly browned surface. The shrimp or filling pieces would probably add \underline{tender}, \underline{springy} bites that contrast with the \underline{smoother} sauce and \underline{soft} base underneath. The baked finish suggests a faint \underline{crust} or \underline{resistance} at the top edges, but most of the dish appears \underline{lush} and \underline{spoonable} rather than crisp. Overall, the texture likely feels \underline{creamy}, \underline{tender}, and substantial, with only light textural variation from the browned surface.\\
\textbf{Sound:} This looks like a fairly \underline{quiet} food to eat, since \underline{creamy baked dishes} rarely produce much \underline{crunch} or \underline{snap}. The only likely audible element would be a faint \underline{scrape} of the utensil against the container and a \underline{soft pull} through the thicker baked top. Any seafood pieces might offer a slight \underline{firm bite}, but not enough to create a strong eating sound. So the sound profile would probably be \underline{muted} and \underline{soft}, dominated more by gentle scooping than by crispness.
\end{tcolorbox}

\begin{tcolorbox}[halign=flush left, colback=cyan!5, colframe=cyan!40, title={\scriptsize\textbf{Gemini~3.1~Pro} --- Taste:4.4\, Smell:4.5\, Texture:4.3\, Sound:3.8}, left=2pt, right=2pt, top=1pt, bottom=1pt, boxrule=0.3pt]
\textbf{Taste:} The heavily \underline{blistered} and \underline{charred} edges of the crust suggest a \underline{smoky}, slightly \underline{bitter} flavor that balances the pie's \underline{richness}. The thick layer of \underline{bright green}, \underline{oily} sauce indicates a \underline{vibrant}, \underline{herbaceous}, and \underline{garlicky} taste profile typical of fresh pesto. The extensive coverage of melted white cheese will provide a \underline{milky}, \underline{salty}, and \underline{savory umami} base that ties the strong flavors together.\\
\textbf{Smell:} The heavily bubbled and \underline{blackened dough} points to a \underline{strong aroma} of \underline{toasted yeast} and \underline{wood-fired smoke}. The generous application of the green sauce likely emits a \underline{pungent}, \underline{fragrant} scent of fresh basil, crushed garlic, and rich olive oil. The baked, \underline{gooey} cheese will contribute a \underline{warm}, \underline{comforting}, and distinctly \underline{rich dairy fragrance} to the overall bouquet.\\
\textbf{Texture:} The deeply charred and prominent air bubbles on the outer crust promise a \underline{rigid}, \underline{brittle}, and highly \underline{crispy} exterior upon the first bite. This \underline{dry crunch} will quickly give way to a \underline{soft}, \underline{chewy}, and \underline{airy} interior dough network beneath the toppings. The heavy layer of \underline{melted cheese} and \underline{oily pesto} will coat the palate with a \underline{slick}, \underline{gooey mouthfeel} that contrasts sharply with the crust.\\
\textbf{Sound:} Biting into the thick, heavily blistered crust will produce a \underline{loud}, \underline{sharp} \underline{crackling} and \underline{crunching} noise. As chewing continues, the thick and \underline{chewy} interior dough will create a \underline{dense}, \underline{muffled} sound profile. The \underline{wet}, \underline{oily} surface of the melted cheese and pesto might generate a faint \underline{squelching} or \underline{smacking} audio alongside the dominant crust crunch.
\end{tcolorbox}

\begin{tcolorbox}[halign=flush left, colback=red!5, colframe=red!50, title={\scriptsize\textbf{LLaVA} --- Taste:3.5\, Smell:4.0\, Texture:3.0\, Sound:3.0}, left=2pt, right=2pt, top=1pt, bottom=1pt, boxrule=0.3pt]
\textbf{Taste:} The model describes a \underline{rich} \underline{creamy} pasta profile with \underline{seafood} and \underline{herb} components.\\
\textbf{Smell:} \underline{Aroma} is inferred as \underline{buttery} and inviting, with \underline{seafood} and \underline{cooked herb} notes.\\
\textbf{Texture:} Predicted texture is a \underline{soft} pasta base with some \underline{chew} from seafood and sauce thickness.\\
\textbf{Sound:} Expected sound is \underline{modest}, mainly \underline{soft clinking/slurping} with limited \underline{crispness}.
\end{tcolorbox}

\begin{tcolorbox}[halign=flush left, colback=orange!5, colframe=orange!50, title={\scriptsize\textbf{Claude~4.6~Opus} --- Taste:3.6\, Smell:3.3\, Texture:3.4\, Sound:1.8}, left=2pt, right=2pt, top=1pt, bottom=1pt, boxrule=0.3pt]
\textbf{Taste:} This is a packaged takeout container labeled ``Southern Scampi Pasta'' from what appears to be a brand called Slim and Hungry, featuring a baked pasta dish with visible shrimp, a \underline{creamy} cheese sauce, and what looks like diced tomato or red pepper mixed throughout. The flavor would center on a \underline{rich}, \underline{buttery}, \underline{garlicky} scampi sauce coating the pasta, with the shrimp providing a \underline{sweet}, \underline{briny} seafood note that anchors the dish. The visible layer of melted and lightly browned cheese across the top suggests a gratinated finish that would add a \underline{salty}, \underline{savory} depth and a slight \underline{nuttiness} from the browning. Being a pre-packaged, ready-to-eat meal, the flavors may be somewhat \underline{muted} compared to freshly prepared scampi, as reheating and holding tend to dull the brightness of garlic and butter, though the overall combination of creamy pasta, shrimp, and cheese remains an inherently satisfying \underline{comfort food} profile.\\
\textbf{Smell:} Upon opening the clear plastic clamshell lid, the primary \underline{aroma} would be a \underline{warm}, \underline{creamy}, \underline{dairy-heavy} scent from the melted cheese and butter-based scampi sauce that has been baked over the pasta. Garlic would be present as a \underline{background note}, though in a pre-packaged product like this the garlic aroma tends to be \underline{mellowed} and integrated rather than \underline{sharp} and \underline{pungent} as it would be in a freshly sauteed scampi. The shrimp would contribute a \underline{subtle}, slightly \underline{sweet}, \underline{oceanic} undertone that mingles with the cheese but does not dominate the aromatic profile. The overall smell impression would be \underline{comforting} and \underline{familiar}, reminiscent of a baked mac and cheese with seafood additions, though lacking the aromatic \underline{intensity} and \underline{freshness} of a dish plated moments after cooking.\\
\textbf{Texture:} The pasta appears to be a short tubular shape, likely penne or rigatoni, that looks well-cooked and coated in a \underline{thick}, \underline{clinging} cream sauce, suggesting a \underline{soft}, \underline{tender} bite with a \underline{smooth}, almost \underline{velvety} mouthfeel from the sauce. The cheese layer on top shows some light browning and appears to have formed a thin, slightly \underline{firmer} skin across the surface, which would provide a mild contrast to the softer pasta beneath when first broken through. The shrimp visible near the edges appear to be medium-sized and cooked through, and in a baked preparation like this they would likely be somewhat \underline{firm} and slightly \underline{rubbery} rather than the tender, snappy texture of freshly sauteed shrimp. The overall textural experience would be predominantly \underline{soft}, \underline{smooth}, and \underline{homogeneous}, which is characteristic of baked pasta dishes where extended heat and a creamy sauce unify the components into a cohesive, comfort-food consistency.\\
\textbf{Sound:} This baked pasta dish would produce very little \underline{audible sound} during eating, as the \underline{soft}, sauce-coated pasta, tender shrimp, and melted cheese are all \underline{quiet} foods that compress easily under the teeth without \underline{cracking}, \underline{snapping}, or \underline{crunching}. The most noticeable sound would be the initial \underline{snap} and \underline{crinkle} of opening the plastic takeout container and peeling back the lid, which is unrelated to the food itself. A fork pressing into the baked cheese top might produce a very faint \underline{crackling} if the gratinated layer has dried slightly, but this would be barely perceptible. The eating experience would be essentially \underline{silent}, dominated by \underline{soft chewing} and the quiet \underline{squish} of cream-coated pasta, making this one of the quieter dishes on the sensory sound spectrum.
\end{tcolorbox}

\subsection{Ice Cream (0015)}
\label{sec:suppl_icecream}
{\small Human GT: Taste=4.1, Smell=3.3, Texture=3.7, Sound=1.8}

\begin{figure}[H]
\centering
\includegraphics[width=0.6\columnwidth]{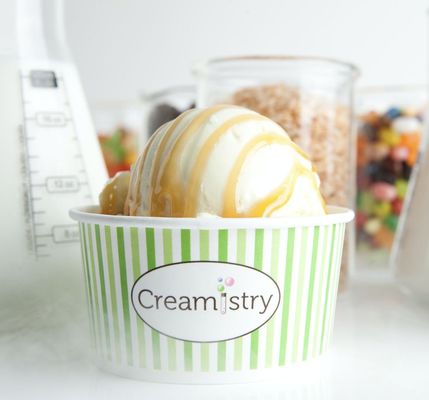}
\end{figure}

\vspace{-6pt}

\scriptsize
\begin{tcolorbox}[halign=flush left, colback=green!5, colframe=green!40, title={\scriptsize\textbf{FoodSense-VL (Stage~2)} --- Taste:4.5\, Smell:4.0\, Texture:4.5\, Sound:3.5}, left=2pt, right=2pt, top=1pt, bottom=1pt, boxrule=0.3pt]
\textbf{Taste:} Visual cues suggest a vanilla-forward \underline{sweet} profile with caramel drizzle adding \underline{buttery} depth.\\
\textbf{Smell:} The dessert likely carries a \underline{sweet dairy aroma} with \underline{caramelized sugar} notes.\\
\textbf{Texture:} The scoop appears \underline{creamy} and \underline{smooth}, with \underline{sticky} syrup contrast from the topping.\\
\textbf{Sound:} Eating sound is likely \underline{subtle}, mostly spoon contact and \underline{soft} melt cues.
\end{tcolorbox}

\begin{tcolorbox}[halign=flush left, colback=blue!5, colframe=blue!40, title={\scriptsize\textbf{Base (Gemma~3 27B)} --- Taste:4.2\, Smell:3.8\, Texture:4.5\, Sound:2.5}, left=2pt, right=2pt, top=1pt, bottom=1pt, boxrule=0.3pt]
\textbf{Taste:} The ice cream appears vanilla-based with caramel drizzle, suggesting \underline{sweet}, \underline{buttery}, and slightly \underline{burnt-sugar} notes. The base likely tastes \underline{creamy} and dairy-rich, with a generally \underline{indulgent} dessert profile.\\
\textbf{Smell:} Visual cues suggest a \underline{sweet}, \underline{milky} \underline{aroma} with prominent \underline{caramel} notes. A \underline{fresh dairy} scent is likely, with possible subtle \underline{vanilla} nuances.\\
\textbf{Texture:} The scoop appears \underline{smooth}, \underline{dense}, and \underline{creamy}, with \underline{melt-in-mouth} behavior. Caramel drizzle looks \underline{viscous} and \underline{sticky}, adding textural contrast.\\
\textbf{Sound:} Sound profile is minimal, mainly spoon scraping and soft contact with the cup. Minor \underline{crunch} might occur only with optional toppings.
\end{tcolorbox}

\begin{tcolorbox}[halign=flush left, colback=gray!5, colframe=gray!50, title={\scriptsize\textbf{InternVL-2.5~26B} --- Taste:4.5\, Smell:4.0\, Texture:4.5\, Sound:3.5}, left=2pt, right=2pt, top=1pt, bottom=1pt, boxrule=0.3pt]
\textbf{Taste:} The ice cream likely offers a \underline{rich}, \underline{creamy} vanilla base, complemented by a \underline{sweet}, caramel drizzle that adds a deeper, \underline{buttery} flavor. The combination of these elements suggests a delightful balance of \underline{sweetness} and \underline{smoothness}.\\
\textbf{Smell:} The \underline{aroma} of the ice cream is likely to be enticing, with the vanilla providing a \underline{warm}, comforting scent and the caramel adding a hint of \underline{toasted sugar}, creating an inviting and appetizing fragrance.\\
\textbf{Texture:} The ice cream itself is probably \underline{smooth} and \underline{velvety}, with a \underline{soft}, \underline{creamy} consistency that \underline{melts} slightly in the mouth. The caramel drizzle adds a contrasting texture, being slightly thicker and more \underline{viscous}, enhancing the overall eating experience.\\
\textbf{Sound:} The sound of scooping the ice cream might be satisfying, with a \underline{soft}, \underline{creamy crunch} as the spoon cuts through the frozen treat. The drizzle of caramel might add a slight glistening sound as it flows over the ice cream, contributing to the sensory experience.
\end{tcolorbox}

\begin{tcolorbox}[halign=flush left, colback=purple!5, colframe=purple!50, title={\scriptsize\textbf{Qwen2.5-VL~32B} --- Taste:4.5\, Smell:4.0\, Texture:4.5\, Sound:2.5}, left=2pt, right=2pt, top=1pt, bottom=1pt, boxrule=0.3pt]
\textbf{Taste:} The image shows a scoop of ice cream in a striped cup, topped with what appears to be a drizzle of caramel sauce. The \underline{creamy} texture of the ice cream suggests a \underline{smooth} and \underline{rich} flavor profile, likely \underline{sweet} and slightly \underline{tangy} from the vanilla base. The caramel drizzle adds a \underline{warm}, \underline{buttery sweetness} that complements the coolness of the ice cream, creating a \underline{balanced} and \underline{indulgent} taste experience.\\
\textbf{Smell:} While the image does not directly convey smell, the visual cues suggest a delightful \underline{aroma}. The vanilla ice cream is likely to have a classic, \underline{sweet}, and slightly \underline{floral} scent, while the caramel drizzle would add a \underline{warm}, \underline{nutty}, and slightly \underline{toasted} fragrance. These scents combined would evoke a comforting and inviting olfactory experience, enhancing the overall sensory appeal.\\
\textbf{Texture:} The ice cream appears \underline{soft} and \underline{creamy}, indicating a \underline{smooth} and \underline{velvety} texture that \underline{melts} gently in the mouth. The caramel drizzle on top likely has a \underline{glossy}, slightly \underline{sticky} texture that contrasts with the cool, \underline{fluffy} consistency of the ice cream. This combination offers a satisfying interplay between the \underline{soft}, \underline{melting} ice cream and the slightly \underline{firmer} caramel topping.\\
\textbf{Sound:} The image itself does not provide auditory information, but one can imagine the sound of the ice cream being scooped into the cup or the slight \underline{crunch} of any potential toppings like nuts or sprinkles. However, the primary focus is on the visual elements, leaving the auditory experience to the imagination. The lack of direct audio cues limits the assessment in this area.
\end{tcolorbox}

\begin{tcolorbox}[halign=flush left, colback=black!5, colframe=black!40, title={\scriptsize\textbf{GPT~5.4~Thinking} --- Taste:4.3\, Smell:3.8\, Texture:4.6\, Sound:2.1}, left=2pt, right=2pt, top=1pt, bottom=1pt, boxrule=0.3pt]
\textbf{Taste:} This dessert looks distinctly \underline{sweet}, \underline{creamy}, and \underline{smooth}, with the pale ice cream and caramel drizzle suggesting a \underline{rich}, \underline{vanilla-forward} flavor. The glossy sauce likely adds \underline{buttery sweetness} and a slightly deeper \underline{toasted-sugar} note that makes the cup feel more \underline{indulgent}. Visually, there are no \underline{sharp} or \underline{bitter} elements, so the flavor impression is \underline{soft}, \underline{rounded}, and \underline{dessert-like} rather than \underline{complex} or contrasting. Overall, it seems likely to taste \underline{mellow}, \underline{sugary}, and \underline{comforting}, with a clean \underline{dairy richness}.\\
\textbf{Smell:} The image suggests a \underline{gentle} but \underline{pleasant aroma} led by \underline{cold dairy}, \underline{vanilla}, and \underline{sweet caramel}. Ice cream usually has a \underline{softer scent} than hot foods, so the smell would likely be noticeable only at \underline{close range} rather than strongly filling the air. The caramel topping implies a faint \underline{buttery}, \underline{cooked-sugar fragrance} that adds \underline{warmth} to the otherwise cool profile. Altogether, it likely smells \underline{sweet} and \underline{inviting}, but relatively \underline{delicate}.\\
\textbf{Texture:} The scoop looks exceptionally \underline{smooth} and \underline{dense}, suggesting a \underline{creamy}, \underline{velvety mouthfeel} with very little \underline{iciness}. The caramel drizzle appears slightly \underline{syrupy}, which would add a \underline{glossy}, \underline{sticky} layer over the colder, \underline{firmer} ice cream surface. As it softens, the dessert would probably become even \underline{silkier} and more \underline{luscious}, coating the mouth evenly. Overall, the texture seems dominated by \underline{softness}, \underline{creaminess}, and a \underline{slow melt}.\\
\textbf{Sound:} This looks like a very \underline{quiet food}, since smooth ice cream and syrup produce almost no natural \underline{crunch} or \underline{crackle}. The main sounds would likely come from the spoon \underline{scraping lightly} against the paper cup and the \underline{soft cut} through the \underline{dense frozen scoop}. Without visible toppings like nuts or crisp mix-ins, there is little reason to expect \underline{sharp biting sounds}. So the sound profile would probably be minimal, \underline{muted}, and \underline{gentle}.
\end{tcolorbox}

\begin{tcolorbox}[halign=flush left, colback=cyan!5, colframe=cyan!40, title={\scriptsize\textbf{Gemini~3.1~Pro} --- Taste:4.5\, Smell:4.6\, Texture:4.3\, Sound:2.5}, left=2pt, right=2pt, top=1pt, bottom=1pt, boxrule=0.3pt]
\textbf{Taste:} The dense, dark brown gravy enveloping the meat suggests a deeply \underline{rich}, \underline{savory}, and heavily \underline{seasoned} flavor profile typical of a robust curry or stew. The large chunks of slow-cooked meat indicate a \underline{meaty}, concentrated \underline{umami} taste that permeates the thick sauce. An underlying \underline{earthy}, \underline{pungent} taste is likely due to the heavily caramelized base, providing a \underline{complex} and possibly \underline{spicy} finish.\\
\textbf{Smell:} The dark, \underline{glossy}, and heavily simmered sauce promises an \underline{intense}, \underline{fragrant aroma} dominated by \underline{toasted spices}, \underline{caramelized onions}, and \underline{roasted meat}. The thick consistency suggests the smell is \underline{heavy} and \underline{lingering}, characteristic of a slow-cooked, \underline{complex} curry. A \underline{robust}, \underline{earthy}, and \underline{pungent} scent will strongly emanate from the hot, deeply colored gravy.\\
\textbf{Texture:} The meat, clearly broken down and coated entirely in the thick sauce, promises an extremely \underline{tender}, yielding, and \underline{melt-in-the-mouth} consistency. The dense, \underline{glossy} gravy will provide a \underline{heavy}, \underline{clinging}, and \underline{velvety} mouthfeel that coats the palate completely. The lack of distinct, separate vegetables suggests a mostly \underline{uniform}, \underline{soft}, and \underline{unctuous} experience, heavily weighted toward the rich sauce.\\
\textbf{Sound:} The \underline{dense} and \underline{wet} nature of the dish implies that interacting with it will produce \underline{heavy}, \underline{dull sloshing} and \underline{soft squelching} noises. Because the meat and sauce are both \underline{soft} and slow-cooked, there will be no \underline{sharp crunch} or \underline{crackle} during eating. Instead, the sound profile will consist of \underline{muted}, \underline{wet slaps} as the thick gravy moves, and \underline{silent}, \underline{tender} yielding during chewing.
\end{tcolorbox}

\begin{tcolorbox}[halign=flush left, colback=red!5, colframe=red!50, title={\scriptsize\textbf{LLaVA} --- Taste:4.5\, Smell:4.0\, Texture:4.0\, Sound:3.5}, left=2pt, right=2pt, top=1pt, bottom=1pt, boxrule=0.3pt]
\textbf{Taste:} The output suggests a \underline{rich} \underline{sweet-cream} base with caramel-driven \underline{buttery sweetness}.\\
\textbf{Smell:} \underline{Aroma} is inferred as \underline{warm} and dessert-like from vanilla-dairy and caramel notes.\\
\textbf{Texture:} Predicted texture is \underline{soft} and \underline{creamy} with \underline{sticky} syrup contrast.\\
\textbf{Sound:} Sound is \underline{low-intensity}, mostly spoon and \underline{soft mouthfeel}-related cues.
\end{tcolorbox}

\begin{tcolorbox}[halign=flush left, colback=orange!5, colframe=orange!50, title={\scriptsize\textbf{Claude~4.6~Opus} --- Taste:4.0\, Smell:3.5\, Texture:4.6\, Sound:1.5}, left=2pt, right=2pt, top=1pt, bottom=1pt, boxrule=0.3pt]
\textbf{Taste:} This is a serving of ice cream from Creamistry, a chain known for making ice cream to order using liquid nitrogen, presented as a generous scoop of what appears to be \underline{vanilla} or \underline{sweet cream} ice cream drizzled with a \underline{golden caramel} or \underline{salted caramel} sauce in a branded paper cup. The base flavor would be a \underline{clean}, \underline{rich}, \underline{creamy} vanilla with a high dairy fat content, as liquid nitrogen freezing produces an exceptionally \underline{smooth} and \underline{dense} product with minimal ice crystal formation. The caramel drizzle adds a warm, \underline{buttery sweetness} with deep \underline{toasted sugar} notes and possibly a hint of \underline{sea salt} if it is the salted variety, creating a pleasant sweet-on-sweet layering that avoids monotony through the contrast of cool cream and warm caramel flavor tones. Jars of toppings visible in the background, including what appear to be granola, jellybeans, and candies, suggest an array of mix-in options, though this particular serving looks elegantly simple and lets the quality of the base ice cream and caramel shine.\\
\textbf{Smell:} The \underline{aroma} of this ice cream would be \underline{delicate} and \underline{subdued} due to the extremely cold temperature, which suppresses volatile compound release, but as the surface begins to temper slightly a gentle, \underline{sweet}, \underline{dairy-rich vanilla} scent would emerge. The caramel sauce, being warmer and more viscous than the frozen base, would be the more aromatically active component, releasing a \underline{toasty}, \underline{buttery}, \underline{cooked-sugar} fragrance with hints of \underline{brown butter} and possibly a faint \underline{salted} note. Because Creamistry uses liquid nitrogen flash-freezing, the ice cream tends to be served at a colder-than-average temperature, which further dampens the olfactory experience in the initial moments compared to traditionally churned ice cream. As the scoop sits and the surface layer begins to soften, the combined vanilla-caramel aroma would gradually intensify into a \underline{warm}, \underline{dessert-shop sweetness}.\\
\textbf{Texture:} Liquid nitrogen ice cream is renowned for its exceptionally \underline{smooth}, \underline{dense}, and \underline{velvety} texture, and this scoop displays the hallmark characteristics of that process with a tight, glossy surface that shows no visible ice crystals or air pockets. The first spoonful would glide across the tongue with an almost \underline{silken} quality, melting evenly and coating the palate with a \underline{rich}, \underline{heavy cream} sensation that feels noticeably \underline{denser} than conventional soft serve or store-bought ice cream. The caramel drizzle adds a contrasting textural element, introducing a slightly \underline{sticky}, \underline{viscous} ribbon that clings to the spoon and stretches as it is pulled, creating a pleasant interplay between the smooth frozen cream and the syrupy sauce. The paper cup itself provides a slight \underline{tactile warmth} in contrast to the frozen contents, and the rounded scoop shape suggests it was portioned with care, maintaining structural integrity that indicates a well-balanced fat and sugar ratio in the base.\\
\textbf{Sound:} Ice cream is among the \underline{quietest} foods to eat, and this serving would produce almost no \underline{audible sound} during consumption, as the \underline{dense}, \underline{creamy} base simply compresses and melts under the pressure of a spoon or against the warmth of the mouth. The most prominent sound would be the spoon scraping lightly against the bottom and sides of the paper cup as the serving is eaten, producing a \underline{soft}, \underline{muffled scratching} that is characteristic of disposable dessert containers. If any of the topping options from the background jars were added, such as granola or candy pieces, they would introduce \underline{crunching} and \underline{crackling} sounds, but as presented this plain scoop with caramel is essentially \underline{silent}. The only other incidental sound might be a faint \underline{sticky separation} noise as the caramel sauce pulls apart between the spoon and the ice cream surface.
\end{tcolorbox}

\begin{tcolorbox}[halign=flush left, colback=brown!5, colframe=brown!50, title={\scriptsize\textbf{Food-Llama~11B} --- Taste:3.0\, Smell:2.0\, Texture:4.0\, Sound:1.0}, left=2pt, right=2pt, top=1pt, bottom=1pt, boxrule=0.3pt]
\textbf{Taste:} The generated output indicates \underline{sweet} vanilla-like ice cream with caramel \underline{richness} and an overall comfort-dessert profile.\\
\textbf{Smell:} \underline{Aroma} is described as \underline{mild} \underline{sweet dairy} with caramel/toffee-like notes.\\
\textbf{Texture:} Texture is described as \underline{smooth} and \underline{creamy} with a \underline{sticky} syrup contrast.\\
\textbf{Sound:} Predicted sound is very \underline{soft}, mostly spoon scraping and gentle \underline{melt/slurp} cues.
\end{tcolorbox}
\clearpage
\begin{table*}[!t]
\centering
\caption{Sensory ratings on six representative food images (1--5 scale). Human GT (mean of $\sim$25 annotators) is shown for comparison. Best model prediction (closest to GT) per cell in \textbf{bold}. ``--'' = unavailable.}
\label{tab:qualitative_comparison}
\footnotesize
\setlength{\tabcolsep}{3pt}
\resizebox{\textwidth}{!}{%
\begin{tabular}{@{}l cccc cccc cccc cccc cccc cccc@{}}
\toprule
& \multicolumn{4}{c}{Sandwich} & \multicolumn{4}{c}{Steak Rice} & \multicolumn{4}{c}{Taco} & \multicolumn{4}{c}{Scampi} & \multicolumn{4}{c}{Ice Cream} \\
\cmidrule(lr){2-5} \cmidrule(lr){6-9} \cmidrule(lr){10-13} \cmidrule(lr){14-17} \cmidrule(lr){18-21}
& T & S & X & U & T & S & X & U & T & S & X & U & T & S & X & U & T & S & X & U \\
\midrule
Human GT & 4.1 & 4.1 & 4.3 & 3.6 & 4.3 & 4.3 & 4.4 & 4.1 & 4.5 & 4.4 & 4.5 & 3.7 & 3.7 & 3.5 & 3.5 & 2.7 & 4.1 & 3.3 & 3.7 & 1.8 \\
\midrule
\multicolumn{21}{l}{\textit{FoodSense-VL}} \\
FoodSense-VL (Stage~2) & 3.5 & 3.0 & 2.5 & 2.0 & 4.0 & 3.5 & 3.0 & 2.5 & 4.2 & 4.0 & 4.5 & 3.8 & 3.5 & 3.0 & 2.5 & 2.0 & 4.5 & 4.0 & 4.5 & 3.5 \\
Base (Gemma~3 27B) & 3.5 & 3.0 & 2.5 & 1.5 & 3.8 & 3.5 & 3.2 & 2.5 & 4.2 & 3.8 & 4.5 & 2.5 & 3.8 & 3.5 & 3.2 & 2.0 & 4.2 & 3.8 & 4.5 & 2.5 \\
\midrule
\multicolumn{21}{l}{\textit{Open-source baselines}} \\
InternVL-2.5~26B & 4.5 & 4.0 & 4.0 & 3.5 & 4.5 & 4.0 & 4.0 & 3.5 & 4.5 & 4.0 & 4.5 & 3.5 & 4.5 & 4.0 & 4.0 & 3.5 & 4.5 & 4.0 & 4.5 & 3.5 \\
Qwen2.5-VL~32B & 4.5 & 4.0 & 4.5 & 2.5 & -- & -- & -- & -- & 4.5 & 4.0 & 4.5 & 2.5 & 4.0 & 4.0 & 4.5 & 2.5 & 4.5 & 4.0 & 4.5 & 2.5 \\
LLaVA & 3.5 & 4.0 & 3.0 & 3.0 & 3.5 & 4.0 & 3.0 & 3.0 & 3.5 & 3.0 & 4.0 & 3.0 & 3.5 & 4.0 & 3.0 & 3.0 & 4.5 & 4.0 & 4.0 & 3.5 \\
Food-Llama~11B & -- & -- & -- & -- & 4.0 & 4.0 & 4.0 & 3.0 & 3.0 & 2.5 & 4.5 & 2.0 & -- & -- & -- & -- & 3.0 & 2.0 & 4.0 & 1.0 \\
\midrule
\multicolumn{21}{l}{\textit{Commercial models}} \\
GPT~5.4~Thinking & 4.2 & 3.9 & 4.3 & 3.4 & 4.6 & 4.4 & 4.4 & 3.7 & 4.5 & 4.1 & 4.7 & 3.8 & 4.4 & 4.5 & 4.2 & 2.6 & 4.3 & 3.8 & 4.6 & 2.1 \\
Gemini~3.1~Pro & 4.2 & 3.8 & 4.5 & 1.2 & 4.5 & 4.6 & 4.3 & 2.8 & 4.6 & 4.4 & 4.8 & 4.5 & 4.4 & 4.5 & 4.3 & 3.8 & 4.5 & 4.6 & 4.3 & 2.5 \\
Claude~4.6~Opus & 4.2 & 3.9 & 4.0 & 3.5 & 4.5 & 4.3 & 4.2 & 3.2 & 4.4 & 4.1 & 4.5 & 3.0 & 3.6 & 3.3 & 3.4 & 1.8 & 4.0 & 3.5 & 4.6 & 1.5 \\
\bottomrule
\end{tabular}
}
\vspace{6pt}
{\scriptsize T=Taste, S=Smell, X=Texture, U=Sound.}
\end{table*}


\subsection{Taco (0005)}
\label{sec:suppl_taco}
{\small Human GT: Taste=4.5, Smell=4.4, Texture=4.5, Sound=3.7}

\begin{figure}[H]
\centering
\includegraphics[width=0.6\columnwidth]{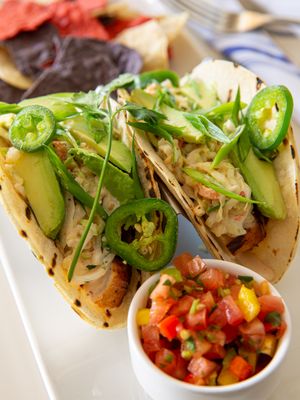}
\end{figure}

\vspace{-6pt}

\scriptsize
\begin{tcolorbox}[halign=flush left, colback=green!5, colframe=green!40, title={\scriptsize\textbf{FoodSense-VL (Stage~2)} --- Taste:4.2\, Smell:4.0\, Texture:4.5\, Sound:3.8}, left=2pt, right=2pt, top=1pt, bottom=1pt, boxrule=0.3pt]
\textbf{Taste:} The tacos show \underline{fresh} avocado, cabbage, pico de gallo, and \underline{grilled} protein, suggesting a balanced \underline{savory} profile with \underline{acidity} and heat. The tortilla color implies lightly \underline{toasted} \underline{nutty} notes.\\
\textbf{Smell:} Steam and \underline{fresh} garnish indicate a \underline{warm}, appetizing \underline{aroma} combining \underline{herbaceous freshness}, \underline{smoky grilled} notes, and \underline{spicy} jalape\~no character.\\
\textbf{Texture:} Strong texture contrast is expected: \underline{soft} avocado, \underline{crisp} cabbage/jalape\~nos, \underline{juicy} pico, and \underline{tender} grilled filling in pliable tortillas.\\
\textbf{Sound:} Likely \underline{moderate} and not loud, with gentle \underline{crunch} from vegetables and subtle \underline{chew} from tortilla and protein.
\end{tcolorbox}

\begin{tcolorbox}[halign=flush left, colback=blue!5, colframe=blue!40, title={\scriptsize\textbf{Base (Gemma~3 27B)} --- Taste:4.2\, Smell:3.8\, Texture:4.5\, Sound:2.5}, left=2pt, right=2pt, top=1pt, bottom=1pt, boxrule=0.3pt]
\textbf{Taste:} The tacos appear to contain grilled fish, likely mild white fish, with slightly \underline{sweet} seafood notes. Avocado and pico suggest \underline{fresh acidity} and \underline{creamy richness}. Jalape\~nos add moderate \underline{heat}, while slaw likely contributes \underline{tangy} balance.\\
\textbf{Smell:} Expected \underline{aroma} combines \underline{smoky}, \underline{savory} grilled fish with \underline{fresh}, \underline{herbaceous} cilantro and \underline{zesty} pico. Avocado adds subtle \underline{buttery} notes and jalape\~nos add a \underline{pungent}, \underline{spicy} edge.\\
\textbf{Texture:} The dish shows strong texture variety: \underline{soft} tortilla, \underline{flaky} fish, \underline{creamy} avocado, \underline{crisp} slaw, and firm jalape\~no slices. Pico appears \underline{juicy} and slightly \underline{chunky}, giving a \underline{complex} mouthfeel.\\
\textbf{Sound:} Eating is likely relatively \underline{quiet}; soft tortilla and fish produce little noise, while slaw and jalape\~nos add occasional light \underline{crunch}. Overall \underline{auditory intensity} remains low.
\end{tcolorbox}

\begin{tcolorbox}[halign=flush left, colback=gray!5, colframe=gray!50, title={\scriptsize\textbf{InternVL-2.5~26B} --- Taste:4.5\, Smell:4.0\, Texture:4.5\, Sound:3.5}, left=2pt, right=2pt, top=1pt, bottom=1pt, boxrule=0.3pt]
\textbf{Taste:} The tacos likely offer a delightful mix of flavors, with the \underline{creamy} coleslaw providing a \underline{tangy} and slightly \underline{sweet} taste, complemented by the \underline{fresh} avocado's \underline{rich}, \underline{buttery} flavor. The jalapenos add a \underline{spicy} kick, while the \underline{grilled} meat offers a \underline{savory}, \underline{smoky} element. The salsa adds a burst of \underline{fresh}, slightly \underline{acidic} flavor, enhancing the overall taste experience.\\
\textbf{Smell:} The \underline{aroma} of the \underline{grilled} meat and tortillas would be prominent, with a \underline{smoky}, slightly \underline{charred} scent. The \underline{fresh} cilantro and avocado would add a \underline{bright}, \underline{herbaceous} note, while the jalapenos would contribute a \underline{mild}, \underline{spicy aroma}. The salsa would add a hint of \underline{fresh}, ripe tomatoes and onions.\\
\textbf{Texture:} The tacos would offer a variety of textures, from the \underline{soft}, slightly \underline{chewy} tortillas to the \underline{crisp}, \underline{fresh} jalapenos and avocado slices. The coleslaw would provide a \underline{crunchy} contrast, while the grilled meat would be \underline{tender} yet slightly \underline{firm}. The salsa would add a \underline{juicy}, \underline{chunky} texture, enhancing the overall \underline{mouthfeel}.\\
\textbf{Sound:} The sound of biting into the tacos would likely be satisfying, with the \underline{crunch} of the jalapenos and avocado, the slight \underline{snap} of the coleslaw, and the \underline{soft} yield of the grilled meat. The salsa would add a \underline{juicy}, slightly messy element, contributing to the auditory experience of eating.
\end{tcolorbox}

\begin{tcolorbox}[halign=flush left, colback=purple!5, colframe=purple!50, title={\scriptsize\textbf{Qwen2.5-VL~32B} --- Taste:4.5\, Smell:4.0\, Texture:4.5\, Sound:2.5}, left=2pt, right=2pt, top=1pt, bottom=1pt, boxrule=0.3pt]
\textbf{Taste:} The tacos appear to combine \underline{grilled} protein, avocado, jalape\~nos, and \underline{fresh} salsa for a \underline{savory} base with \underline{creamy} and \underline{spicy} contrast.\\
\textbf{Smell:} Visual cues suggest a \underline{fresh} \underline{aromatic} profile with herbs/vegetables, \underline{mild smoke} from grilled filling, and \underline{zesty} salsa character.\\
\textbf{Texture:} \underline{Soft} tortilla and avocado contrast with \underline{crunch} from jalape\~nos and \underline{chunky} salsa; overall \underline{mouthfeel} appears varied and engaging.\\
\textbf{Sound:} The expected sound profile is \underline{moderate}, with occasional \underline{crunch} from \underline{crisp} components and otherwise \underline{soft} bite sounds.
\end{tcolorbox}

\begin{tcolorbox}[halign=flush left, colback=black!5, colframe=black!40, title={\scriptsize\textbf{GPT~5.4~Thinking} --- Taste:4.5\, Smell:4.1\, Texture:4.7\, Sound:3.8}, left=2pt, right=2pt, top=1pt, bottom=1pt, boxrule=0.3pt]
\textbf{Taste:} These tacos look \underline{bright}, \underline{savory}, and \underline{fresh}, with the avocado, cabbage slaw, jalape\~nos, and salsa suggesting a balance of \underline{richness}, \underline{acidity}, and \underline{heat}. The \underline{grilled tortillas} and likely seasoned filling underneath imply a \underline{warm}, \underline{smoky} base flavor that would support the cooler toppings. The green jalape\~no slices indicate a noticeable \underline{spicy} kick, while the pico de gallo adds \underline{juicy sweetness} and \underline{tang} from tomato and citrus. Overall, the flavor seems \underline{layered} and \underline{lively}, combining \underline{fresh vegetal notes} with \underline{creamy}, \underline{spicy}, and \underline{grilled} elements.\\
\textbf{Smell:} The strongest likely aromas come from the \underline{warmed corn tortillas} and the cooked filling, which suggest a \underline{toasty}, \underline{savory scent} rising from the plate. Fresh jalape\~no, herbs, and salsa would add \underline{sharp green notes} and a slightly \underline{citrusy brightness} that makes the dish smell \underline{vivid} and \underline{clean}. The avocado contributes less aroma visually, but it reinforces the impression of \underline{freshness} and \underline{richness}. Altogether, the smell likely feels \underline{inviting} and \underline{energetic}, with a mix of \underline{grilled warmth} and \underline{crisp produce}.\\
\textbf{Texture:} The tacos appear especially \underline{textural}, with \underline{soft} avocado, \underline{crunchy} shredded cabbage, \underline{tender} filling, and lightly \underline{charred} tortillas all stacked together. The jalape\~no slices would add a \underline{crisp snap}, while the salsa looks \underline{juicy} and \underline{chunky} enough to bring bursts of moisture. The tortillas seem \underline{pliable} but slightly \underline{blistered}, so they would probably fold softly while still offering a faint \underline{toasted resistance}. Overall, the eating experience looks highly varied and satisfying, moving between \underline{creamy}, \underline{crisp}, \underline{tender}, and \underline{juicy} textures in each bite.\\
\textbf{Sound:} These tacos would likely make a \underline{modest} but \underline{noticeable} eating sound, mostly from the \underline{crunch} of cabbage and the \underline{snap} of fresh jalape\~no. The tortillas seem \underline{soft} enough to avoid a \underline{brittle crack}, though their \underline{charred edges} may give a slight \underline{dry bite sound}. The \underline{juicy filling} and avocado would \underline{dampen} louder sounds quickly, keeping the overall profile \underline{softer} than crunchy fried food. So the sound would probably be a mix of \underline{light crispness} and \underline{moist chewing} rather than anything loud or sharp.
\end{tcolorbox}

\begin{tcolorbox}[halign=flush left, colback=cyan!5, colframe=cyan!40, title={\scriptsize\textbf{Gemini~3.1~Pro} --- Taste:4.6\, Smell:4.4\, Texture:4.8\, Sound:4.5}, left=2pt, right=2pt, top=1pt, bottom=1pt, boxrule=0.3pt]
\textbf{Taste:} The heavily \underline{browned}, \underline{craggy} exterior of the fried chicken suggests a deeply \underline{savory} and \underline{rich} flavor profile, enhanced by the fat of the frying oil. The \underline{shiny}, \underline{clinging sauce} coating the pieces points toward a \underline{robust}, \underline{salty}, and likely \underline{tangy} or \underline{sweet glaze} that dominates the palate. A generous dusting of grated hard cheese and fresh green herbs will provide an additional layer of \underline{sharp}, \underline{salty umami} and a fleeting, \underline{bright herbal finish} to balance the richness.\\
\textbf{Smell:} The visible \underline{thick}, \underline{crispy breading} and \underline{slick sauce} strongly imply an \underline{intense}, \underline{fragrant aroma} of hot frying oil and \underline{savory roasted meat}. The glossy glaze likely emits \underline{pungent}, mouth-watering scents of garlic, vinegar, or caramelized sugars depending on the specific sauce profile. Meanwhile, the white dusting of grated cheese on top will introduce a distinctly \underline{sharp}, \underline{pungent dairy note} that cuts through the heavy fried aromas.\\
\textbf{Texture:} The extremely \underline{jagged} and \underline{blistered crust} enveloping the chicken pieces promises a \underline{rigid}, \underline{brittle}, and highly \underline{crunchy} initial bite. This substantial crunch is immediately contrasted by the \underline{slick}, \underline{sticky mouthfeel} of the wet sauce heavily painted across the surface. Beneath this thick exterior armor, the \underline{plump} pieces of poultry imply a \underline{tender}, \underline{fibrous}, and \underline{juicy} interior that easily yields to chewing.\\
\textbf{Sound:} Piercing the deeply \underline{rigid} and \underline{uneven fried coating} will produce a \underline{loud}, \underline{sharp cracking} and \underline{shattering sound} upon the first bite. Continuous chewing will yield a sustained, \underline{noisy crunching} audio profile as the thick crust breaks down in the mouth. Additionally, the \underline{heavy}, \underline{wet glaze} might generate faint \underline{smacking} or \underline{squelching noises} that complement the dominant, aggressive crunch of the breading.
\end{tcolorbox}

\begin{tcolorbox}[halign=flush left, colback=orange!5, colframe=orange!50, title={\scriptsize\textbf{Claude~4.6~Opus} --- Taste:4.4\, Smell:4.1\, Texture:4.5\, Sound:3.0}, left=2pt, right=2pt, top=1pt, bottom=1pt, boxrule=0.3pt]
\textbf{Taste:} These appear to be fish or chicken tacos in \underline{charred corn tortillas}, generously topped with sliced avocado, fresh jalape\~no rounds, shredded cabbage slaw, and green onion, with a side of fresh pico de gallo containing diced tomato, onion, and what looks like mango or yellow pepper. The flavor profile would be remarkably \underline{layered}, starting with a \underline{mild}, \underline{savory} protein base and building through the \underline{creamy richness} of the avocado, the \underline{bright heat} of the raw jalape\~no slices, and the \underline{tangy crunch} of the cabbage slaw which likely has a lime or vinegar dressing. The pico de gallo would deliver a burst of \underline{fresh acidity} and \underline{sweetness} that cuts through the richer elements, while the lightly charred corn tortillas contribute a \underline{subtle smoky}, \underline{earthy corn flavor}. Blue corn tortilla chips visible in the background along with what appears to be refried black beans would add a \underline{salty}, \underline{starchy} complement to round out this \underline{vibrant}, \underline{well-balanced} meal.\\
\textbf{Smell:} The aroma would be led by the \underline{fresh}, \underline{green}, \underline{herbaceous} scent of the raw jalape\~no slices and green onions piled on top, which release \underline{sharp}, \underline{peppery} volatile compounds even before being eaten. The charred corn tortillas would contribute a \underline{warm}, \underline{toasty}, slightly \underline{smoky} fragrance that serves as an \underline{earthy} backdrop to the brighter toppings. The pico de gallo in the small white ramekin would emit a \underline{clean}, \underline{garden-fresh} scent of ripe tomato, raw onion, and likely cilantro and lime juice, creating an invigorating \underline{citrus-herbal} note. The avocado itself is relatively mild in aroma but adds a \underline{subtle}, \underline{fatty}, slightly \underline{grassy undertone} that blends with the overall \underline{fresh} and \underline{lively} olfactory character of the dish.\\
\textbf{Texture:} The textural complexity of these tacos is exceptional, beginning with the \underline{soft} yet slightly \underline{chewy} corn tortillas that show visible char blistering, giving them a \underline{pliable} structure with occasional \underline{crisp spots} along the scorched edges. The shredded cabbage slaw provides a \underline{crunchy}, \underline{juicy}, and \underline{springy} bite that is the primary textural counterpoint to the softer ingredients, while the avocado slices contribute a \underline{buttery}, \underline{smooth}, almost \underline{silky} mouthfeel that melts against the palate. The jalape\~no rounds would add a \underline{firm}, \underline{crisp}, and \underline{watery snap} with each bite, and the protein underneath appears shredded or pulled, suggesting a \underline{tender}, \underline{flaky} consistency. The pico de gallo on the side offers a \underline{chunky}, \underline{juicy}, and slightly \underline{pulpy} texture that would add further variety when spooned onto each bite.\\
\textbf{Sound:} Biting into these tacos would produce a \underline{moderate initial crunch} from the charred edges of the corn tortilla and the raw cabbage slaw, though the overall sound profile is relatively \underline{subdued} given the predominance of \underline{soft}, \underline{moist} ingredients. The fresh jalape\~no slices would contribute a clean, \underline{crisp snap} when bitten through, briefly punctuating the \underline{quieter chewing} of avocado and shredded protein beneath. Scooping pico de gallo with a blue corn chip from the side would generate a \underline{satisfying}, \underline{sharp cracking sound} as the chip breaks, which would be the loudest acoustic element of this meal. The overall eating experience leans toward \underline{quiet} and \underline{soft}, with intermittent \underline{crunchy} moments from the slaw, jalape\~nos, and chips providing textural and auditory contrast.
\end{tcolorbox}

\begin{tcolorbox}[halign=flush left, colback=red!5, colframe=red!50, title={\scriptsize\textbf{LLaVA} --- Taste:3.5\, Smell:3.0\, Texture:4.0\, Sound:3.0}, left=2pt, right=2pt, top=1pt, bottom=1pt, boxrule=0.3pt]
\textbf{Taste:} The taco image suggests \underline{savory} meat, \underline{creamy} avocado, and \underline{tangy} salsa with \underline{mild spice} and \underline{freshness} from green toppings.\\
\textbf{Smell:} \underline{Aroma} is inferred as a mix of \underline{savory cooked} filling, \underline{fresh} salsa, and \underline{mild spicy} notes.\\
\textbf{Texture:} Visual cues indicate \underline{soft} tortilla and \underline{creamy} avocado with \underline{crisp} vegetables and \underline{tender} filling for mixed bite texture.\\
\textbf{Sound:} Likely a \underline{moderate} combination of \underline{soft chew} and small \underline{crunch} events from \underline{crispy} elements.
\end{tcolorbox}

\begin{tcolorbox}[halign=flush left, colback=brown!5, colframe=brown!50, title={\scriptsize\textbf{Food-Llama~11B} --- Taste:3.0\, Smell:2.5\, Texture:4.5\, Sound:2.0}, left=2pt, right=2pt, top=1pt, bottom=1pt, boxrule=0.3pt]
\textbf{Taste:} The dish appears to offer familiar taco flavor balance: \underline{savory} filling, \underline{creamy} avocado, and \underline{zesty} salsa in a \underline{soft} tortilla.\\
\textbf{Smell:} Likely \underline{aromatic} but \underline{moderate}, combining \underline{fresh citrus-herb} notes and a \underline{mild cooked-protein} smell.\\
\textbf{Texture:} The output emphasizes strong texture diversity with \underline{soft}, \underline{creamy}, and \underline{crisp} components in one bite.\\
\textbf{Sound:} Sound is expected to be relatively \underline{soft} with occasional \underline{crunch} from \underline{fresh/crisp} toppings.
\end{tcolorbox}

\end{document}